\documentclass[unnumsec,webpdf,contemporary,large]{main}%
\graphicspath{{Fig/}}

\theoremstyle{thmstyleone}%
\theoremstyle{thmstyletwo}%
\theoremstyle{thmstylethree}%

\usepackage{comment}
\usepackage{natbib}
\usepackage{hyperref}
\usepackage{subcaption}
\usepackage[margin=1in]{geometry}
\usepackage{booktabs} % For formal tables
\usepackage{tabularx} % Allows for table cells that wrap text

\begin{document}

\journaltitle{Briefings in Bioinformatics}
\DOI{DOI HERE}
\copyrightyear{2024}
\pubyear{2024}
\access{Advance Access Publication Date: Day Month Year}
\appnotes{Paper}

\firstpage{1}

%\subtitle{Subject Section}

\title[Feature Selection on Knowledge Graphs]{A review of Feature Selection on Knowledge Graphs}
\title[Feature Selection on Knowledge Graphs]{A review of feature selection strategies utilizing graph data structures and knowledge graphs}

\author[1]{Sisi Shao\ORCID{0009-0000-9783-9205}}
\author[2]{Pedro Henrique Ribeiro}
\author[1]{Christina Ramirez\ORCID{0000-0002-8435-0416}}
\author[1,2,$\ast$]{Jason H. Moore\ORCID{0000-0002-5015-1099}}

\authormark{Author Name et al.}

\address[1]{\orgdiv{Department of Biostatistics}, \orgname{Fielding School of Public Health at University of California, Los Angeles}, \orgaddress{\street{650 Charles E Young Dr S}, \postcode{90095-1772}, \state{California}, \country{Country}}}

\address[2]{\orgdiv{Department of Computational Biomedicine}, \orgname{Cedars-Sinai Medical Center}, \orgaddress{\street{8700 Beverly Blvd}, \postcode{90048}, \state{California}, \country{United States}}}

\corresp[$\ast$]{Corresponding author. \href{email:Jason.Moore@csmc.edu}{Jason.Moore@csmc.edu}}

\received{Date}{0}{Year}
\revised{Date}{0}{Year}
\accepted{Date}{0}{Year}

%\editor{Associate Editor: Name}

%\abstract{
%\textbf{Motivation:} .\\
%\textbf{Results:} .\\
%\textbf{Availability:} .\\
%\textbf{Contact:} \href{name@email.com}{name@email.com}\\
%\textbf{Supplementary information:} Supplementary data are available at \textit{Journal Name}
%online.}

\abstract{
Feature selection in Knowledge Graphs (KGs) are increasingly utilized in diverse domains, including biomedical research, Natural Language Processing (NLP), and personalized recommendation systems. This paper delves into the methodologies for feature selection within KGs, emphasizing their roles in enhancing machine learning (ML) model efficacy, hypothesis generation, and interpretability. Through this comprehensive review, we aim to catalyze further innovation in feature selection for KGs, paving the way for more insightful, efficient, and interpretable analytical models across various domains. Our exploration reveals the critical importance of scalability, accuracy, and interpretability in feature selection techniques, advocating for the integration of domain knowledge to refine the selection process. We highlight the burgeoning potential of multi-objective optimization and interdisciplinary collaboration in advancing KG feature selection, underscoring the transformative impact of such methodologies on precision medicine, among other fields. The paper concludes by charting future directions, including the development of scalable, dynamic feature selection algorithms and the integration of explainable AI principles to foster transparency and trust in KG-driven models.
}
 
%\keywords{Feature Selection, Knowledge Graphs, Search Algorithms, Vector Embeddings, Similarity-based Methods, Advanced Network Representation Learning, Deep Learning, Precision Medicine, Complex Diseases, Explainable AI, ML.}

% \boxedtext{
% \begin{itemize}
% \item Key boxed text here.
% \item Key boxed text here.
% \item Key boxed text here.
% \end{itemize}}

\maketitle
 
%\section*{Author summary}

%\linenumbers

% Use "Eq" instead of "Equation" for equation citations.
%\tableofcontents
%\listoffigures

\section{Introduction}
%\section*{Table of Acronym}

\subsection{Brief Introduction to Knowledge Graphs}

In the era of large-scale digital information, Knowledge Graphs (KGs) are an increasingly popular tool for organizing data and information\cite{chicaiza2021comprehensive}. At their core, KGs are characterized by representing entities and their relationships through triplets (subject-predicate-object), allowing for in-depth data analysis and the development of personalized care strategies. For instance, a triplet like ``Cyclophosphamide - treats - Cancer" demonstrates KGs' potential in drug discovery and repurposing. Platforms like Bio2RDF have been instrumental in exploring the complex relationships between genetics, diseases, and environmental factors. KGs thereby facilitate a comprehensive approach to healthcare; this approach supports a wide range of applications, from advanced decision-support systems to personalized medicine and innovative drug discovery methods \citep{belleau2008bio2rdf, hasan2020knowledge}.

One of the most well-known uses for KGs is in the development of web-based technologies, including search engines and the Semantic Web (an extension of the World Wide Web that enables data to be shared and reused across applications). Google KG, DBpedia, and Yet Another Great Ontology (YAGO) utilize the principles of the Semantic Web and Linked Open Data (LOD\textendash a method of publishing structured data so that it can be interlinked and become more useful) to create extensive networks of nodes and edges, representing the intricate relationships within vast datasets and enabling enhanced query processing and analytics capabilities. The contributions of scholars such as Fensel et al. \cite{fensel2020introduction}, Bonner et al. \cite{bonner2022understanding}, and Yang et al. \cite{yang2023comprehensive} have been crucial in shedding light on the foundational aspects and ongoing evolution of these systems.

As biology continues to advance, we're accumulating a vast amount of knowledge about genes, proteins, chemicals, cells, diseases, and other biological entities along with their complex interactions which are intricate and multifaceted\cite{levine2019biological}. To make sense of this complexity, KGs have emerged as a powerful tool for organizing and connecting this information. In the realm of precision medicine, KGs have been used to consolidate disparate biomedical data, and been applied to improving the effectiveness of personalized patient care by systematically utilizing genetic, environmental, and lifestyle information. This application is exemplified by PrimeKG, which significantly contributes to creating a comprehensive medical knowledge base by integrating a wide ontology with data from various sources, including genomic databases, thereby supporting detailed medical research and personalized care planning \citep{chandak2023building}.

At their core, KGs are characterized by representing entities and their relationships through triplets (subject-predicate-object), allowing for in-depth data analysis and the development of personalized care strategies. For instance, a triplet like ``Cyclophosphamide - treats - Cancer" demonstrates KGs' potential in drug discovery and repurposing. Platforms like Bio2RDF have been instrumental in exploring the complex relationships between genetics, diseases, and environmental factors. KGs thereby facilitate a comprehensive approach to healthcare; this approach supports a wide range of applications, from advanced decision-support systems to personalized medicine and innovative drug discovery methods \citep{belleau2008bio2rdf, hasan2020knowledge}.

The integration and analysis of data from biomedical research and clinical practice through KGs provide a dynamic platform for advancements in understanding and treating diseases. The academic discourse on feature selection methods applied to KGs, as highlighted by the studies referenced, underscores their transformative potential in various domains, particularly in advancing personalized medicine and healthcare outcomes.

\subsection{Importance of Feature Selection}
Feature selection involves choosing a subset of input variables most relevant for analysis, crucial in modern ML research due to the vast amounts of data ranging from petabytes to exabytes. As datasets grow in size and complexity, identifying important attributes is essential to address the "curse of dimensionality" \citep{bellman1966dynamic}, which can degrade model performance. Reducing the feature set helps mitigate overfitting and improves computational efficiency \citep{ferreira2012efficient}. This reduction aids model interpretability in critical domains like healthcare and finance \citep{lahmiri2016features, huda2016hybrid} and enhances the model's generalizability to new data, a cornerstone for practical applications \citep{forster2000key, saari2010generalizability}. Streamlined models, requiring fewer computational resources, are beneficial in resource-constrained scenarios like edge computing \citep{bikku2016hadoop, mohammed2020edge}. With big data's growing influence, especially in healthcare projected to reach \$79.23 billion by 2028, feature selection is increasingly vital to ensure robust and applicable models.

Often in ML, feature selection refers to selecting columns of a tabular dataset. In this paper, we take a broader view, including selecting nodes or entities for hypothesis generation and further investigation. For example, a knowledge graph (KG) with genes and diseases can hypothesize new subsets of genes related to a specific disease.

Recognizing various feature selection methods, such as algorithmic techniques, statistical analyses \citep{jovic2015review}, and expert insights, we now explore the relationship between KGs and feature selection, highlighting how these frameworks can enhance the feature selection process.

\begin{figure*}
    \centering
    \includegraphics[scale=0.32]{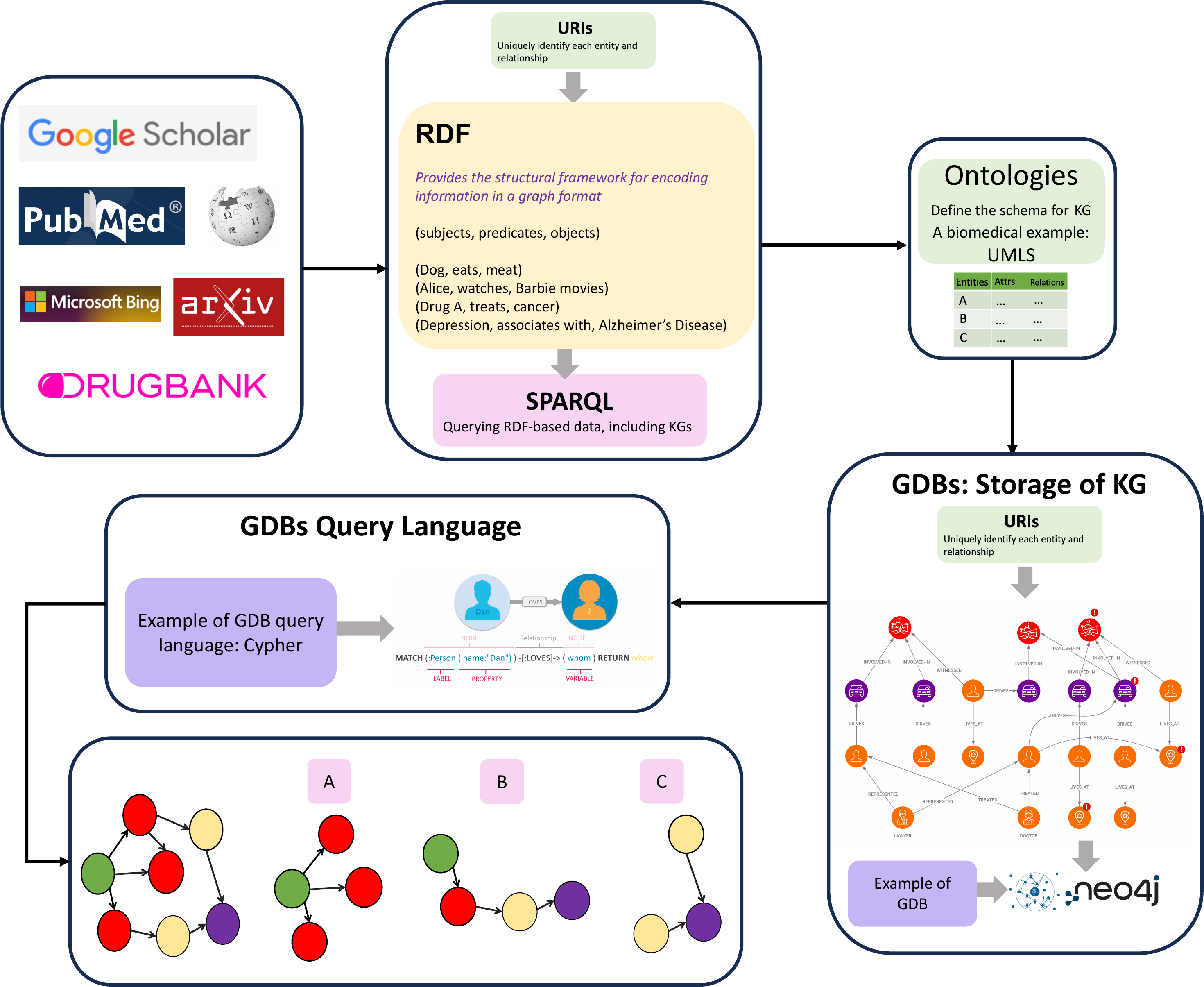}
    \caption{An integrated overview of KGs encompassing RDF structuring, Ontological frameworks, and GDB management, illustrating the flow from data sources to semantic querying and storage. Figure \ref{fig:kg-construction} delineates the contribution of varied scholarly and scientific data sources—such as Google Scholar, PubMed, arXiv, and DrugBank—in providing raw data inputs. These inputs are then semantically encoded via the RDF, using triples that consist of subjects, predicates, and objects, alongside URIs that ensure the unique identification and integration of data entities across the KG. At the heart of the semantic structure are ontologies, exemplified here by the Unified Medical Language System (UMLS), which define the schema for the KG by outlining the essential relationships and attributes of the domain-specific entities. This ontology-based schema informs the organization and representation of knowledge within GDBs, such as Neo4j, which are specialized for storing and operationalizing the complex relational data of KGs. The central round-edged box showcases the role of query languages, with Cypher portrayed as a model for extracting information from GDBs through its intuitive syntax and pattern matching capabilities. The graphic elucidation of the query output illustrates a network of nodes and edges, representing the intricate interrelations and potential analytical insights derived from KGs. Each cluster within the network, designated as A, B, and C, symbolizes distinct subsets or aspects of the graph database that have been queried. }
    \label{fig:kg-construction}
\end{figure*}

\subsection{Overview of the Relationship between Knowledge Graphs and Feature Selection}

The integration of KGs with feature selection processes marks a pivotal advancement in the realm of ML, particularly enhancing the capabilities of predictive models. Notably, many AI/ML systems remain largely unaware of domain-specific knowledge, such as biomedical information, which humans routinely leverage to solve complex problems. This oversight highlights the potential of KGs, with their rich web of entities, attributes, and interconnections, to bridge this gap. KGs play a crucial role across diverse domains such as the Semantic Web, NLP, and comprehensive data integration efforts, providing a structured representation that significantly aids in the precision of feature selection. This critical phase in ML aims at pinpointing the most relevant data attributes to optimize model performance, reduce over-fitting, and enhance interpretability.

% Within the Semantic Web’s framework, KGs serve as the backbone for making web content intelligible to both machines and humans, thereby facilitating sophisticated data retrieval mechanisms and automated processes. Ontologies, integral to KGs, delineate a domain's knowledge through a meticulously structured network of concepts and relationships. This establishes a universal language that ensures data consistency and supports logical inferences from the graph. The symbiotic relationship between KGs and ontologies significantly bolsters the feature selection process by ensuring that selected attributes align with domain-specific semantics and relationships, thereby enhancing the model's discriminative power.

However, integrating KGs into feature selection is challenging, encompassing issues of scalability, KG integrity, and adaptation to diverse domains. These challenges call for a concerted research effort aimed at developing scalable algorithms that efficiently navigate expansive KGs, enhance KG completeness, and foster the integration of varied data sources. This multidisciplinary arena benefits immensely from the combined expertise in knowledge representation, ML, and domain-specific areas, underscoring the critical need for a harmonious blend of structured knowledge with empirical, data-driven approaches.

To this end, exploring innovative methodologies becomes paramount. Approaches such as embedding-based feature selection and the application of graph neural networks (GNNs) demonstrate the potential of leveraging KGs' unique characteristics for feature selection. These methodologies offer scalable and effective solutions for managing the high-dimensional spaces inherent to KGs, thus facilitating a more nuanced and comprehensive analysis of data.

Moreover, the dynamic nature of KGs, with their constantly evolving entities and relationships, necessitates feature selection methods that are not only adaptive but also capable of real-time updates. This adaptability ensures the relevance and efficacy of selected features in the face of new information, thereby maintaining the integrity and applicability of ML models in rapidly changing scenarios.

\section{Background and Key Concepts}
\label{sec:background}
\subsection{Definition and Structure of Knowledge Graphs}
KGs categorize and link data for domain-specific knowledge discovery.

\subsubsection{Ontologies}
KGs use ontologies to define relationships and model semantics \citep{staab2010handbook}. Ontologies categorize concepts, allowing flexible queries. Bio2RDF, for example, defines classes like "proteins" and "chemical entities," and their relationships using RDF triples.

\subsubsection{Example: Bio2RDF}
Bio2RDF integrates datasets like DrugBank \citep{wishart2018drugbank}, SIDER \citep{kuhn2016sider}, and KEGG \citep{kanehisa2000kegg} into a unified RDF structure, enhancing data interoperability and supporting complex queries.

\begin{itemize}
    \item \textbf{Nodes:} Tagged with URIs, representing biomedical entities like genes and drugs.
    \item \textbf{Relationships:} Include "targets" and "is affected by," illustrating drug-protein interactions and genetic influences.
\end{itemize}

\subsection{Structuring Domain Knowledge with RDF}

\subsubsection{RDF}
RDF provides a structure for semantic representation in KGs \citep{bodenreider2004unified}. It formalizes relationships as triplets (subject-predicate-object) forming a graph $G = \{(s, p, o)\}$. RDF enhances data interlinking and queryability \citep{donnelly2006snomed, nelson2011normalized}.

\subsubsection{Ontologies}
Ontologies in KGs categorize and describe concepts with flexible relationships. They enhance querying capabilities by defining both specific and abstract relationships, as seen in Bio2RDF and AlzKB.

\subsection{Leveraging Graph Databases}
Graph Databases (GDBs) like Neo4j manage complex data relationships within KGs, enabling efficient semantic analysis \citep{miller2013graph}. Freebase and query languages like Cypher and SPARQL extend GDB functionality for intuitive querying \citep{bollacker2008freebase, francis2018cypher}.

\subsection{Visual Demonstration of ADKGs of Varying Sizes}
\begin{figure*}[ht]
    \centering
    \includegraphics[scale=0.38]{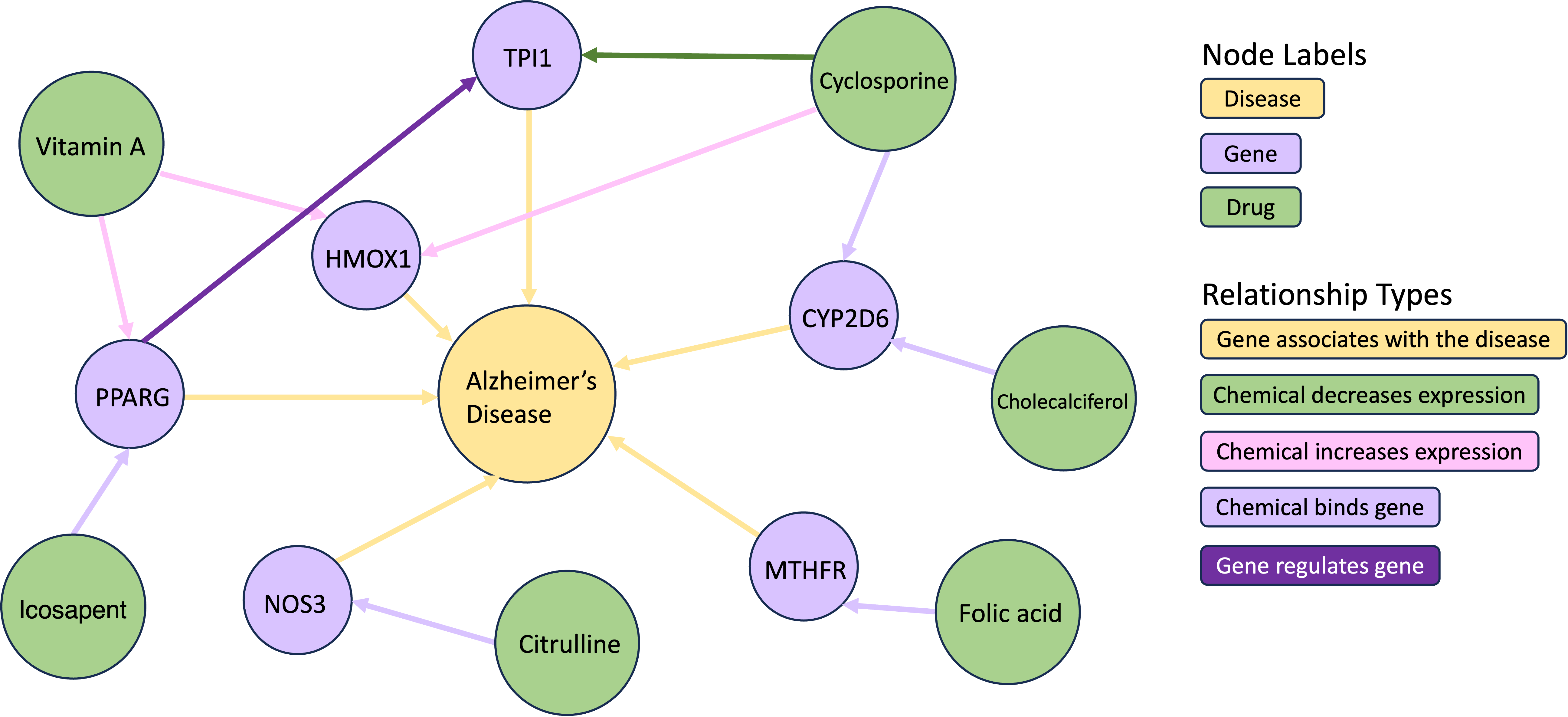}
    \caption{A Tiny-sized ADKG (Yellow Node: AD; Purple Nodes: Genes; Green Nodes: Drugs) \citep{alzheimersknowledgebase}. There are five instances of the ``Chemical binds gene" relationship (light purple arrows), where a chemical is shown to interact directly with a gene; six instances of the ``Gene associates with disease" relationship (yellow arrows), representing genes that have an association with AD; one instance of the ``Chemical decreases expression"
    relationship (dark green arrow), indicating a chemical that downregulates or decreases the expression of a gene; one instance of  ``Gene regulates gene" (purple arrow), suggesting a regulatory interaction between two genes, PPARG and TPI1. More detailed information on genes and drugs is given in the Appendix B. }
    \label{fig:alzkg_tiny}
\end{figure*}

We use AlzKb, an Alzheimer's Disease KG, to demonstrate varying KG sizes. Figures represent tiny (Cypher query limit 8), small (Cypher query limit 15), and medium (CYpher query limit 200) KGs. A tiny KG example is shown in Figure~\ref{fig:alzkg_tiny}.

%\subsubsection{Small KG}

\section{Feature Selection on Knowledge Graphs}
\label{sec:feature_selection_knowledge_graphs}
%\subsection{Review of Algorithms Specific to Knowledge Graphs}
We categorize and evaluate the methodological frameworks delineated within the referenced manuscripts in this section. Below we elaborate on four distinct KG feature selection methods, including search algorithms, similarity-based methods, vector embeddings, and advanced network representation learning, available in the most current literature to the best of our present knowledge.

\subsection{Causal Discovery-Search Algorithm}

The goal of causal discovery is to move beyond merely describing correlated events to identifying the direction of influence between observed phenomena. The challenge in causality analysis lies in capturing the complex interactions between variables. Typically, these relationships are formalized using causal graphs, where nodes represent variables and directed edges denote causal effects.

In medicine, the gold standard for establishing causal relationships, including confounding, collider, mediation, moderation, reverse causality, effect modification, causal chain, and causal graph, is through randomized controlled trials (RCTs). However, various analytical methods can infer causal relationships from observational data. Researchers must consider other measured or unmeasured variables that may act as confounders, mediators, or colliders.  For a comprehensive review of causal discovery, we recommend this survey paper by Zanga et al. \cite{zanga2022survey}.

There has been a lot of work recently on building automated methods, generally utilizing natural language processing (NLP) techniques, to extract causal relations from the scientific literature. These KGs can then be used to consolidate knowledge and form inferences and hypotheses about how different variables interact. Causal analysis can then be used to identify features that have causal effects on downstream variables. 

The study by Malec et al. \cite{malec2023causal} introduced a novel causal feature selection framework using the "ADKG" knowledge graph. This ADKG was constructed from post-2010 PubMed biomedical literature and an ontology-grounded KG via the PheKnowLator workflow \citep{callahan2019pheknowlator}. The authors used PubMed identifiers and machine reading systems like EIDOS, REACH, and SemRep within the INDRA ecosystem \citep{gyori2017word} to extract data. INDRA assembles knowledge into a model of causal molecular interactions \citep{perez2009semantics}, resulting in an OWL ontology \citep{horrocks2005owl}.

The study aimed to enhance causal feature selection with the ADKG. Hygiene steps were performed, and logical entailments were initially omitted. Predicates were mapped to the Relation Ontology (RO) to provide logical definitions and infer additional knowledge. Forward-chaining inference using CLIPS generated new triples based on RO properties, with belief scores assigned. Integration with PheKnowLator facilitated path search algorithms, reweighting edges with hierarchical relationships for optimized path searches. Competency questions, such as causal relationships between depression and AD, were addressed using SPARQL queries and Dijkstra's shortest path algorithm \citep{perez2009semantics, ducharme2013learning}.

\begin{figure}
    \centering
    \includegraphics[scale=0.3]{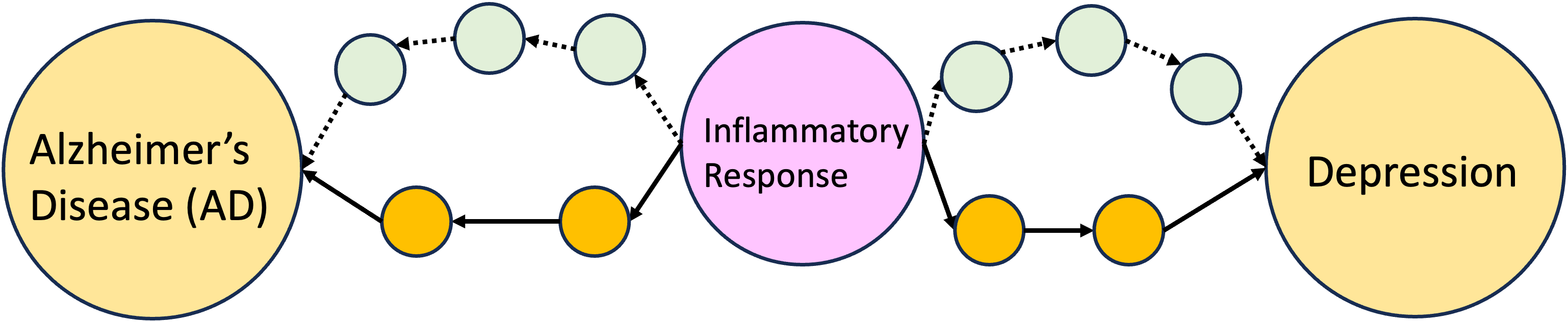}
    \caption{Illustration of Inflammatory Response (pink node) as a Potential Confounder in the Association Between AD (left yellow node) and Depression (right yellow node). The diagram represents the shortest paths (through orange nodes) identified by Dijkstra's algorithm. The two green paths also connect inflammatory response with AD and Depression but both of them are one unit longer than the orange ones. Consequently, Dijkstra's algorithm picks the shortest path.}
    \label{fig:causalfeature}
\end{figure}

Dijkstra's algorithm applied to the ADKG identified shortest paths connecting genes and diseases, highlighting direct relationships \citep{malec2023causal}. These paths were analyzed to identify potential confounders, colliders, and mediators. Confounders influence both exposure and outcome, colliders are influenced by both, and mediators act as intermediaries. Figure~\ref{fig:causalfeature} illustrates identifying a potential confounder between AD and depression using Dijkstra's algorithm. The study identified 126 unique potential confounders, 29 colliders, and 18 potential mediators, showcasing the ADKG's ability to uncover intricate relationships that traditional searches might miss.

\subsection{Feature Selection-Dimensionality Reduction}
KGs can be utilized to perform feature selection for high-dimensional tabular datasets. In this scenario, nodes in the graph may relate to the columns, or features, of the tabular dataset. These subsets of features then be analyzed with methods such as machine learning. Below, we outline a few examples of graph-based methods for selecting subsets of features.  
\begin{itemize}
    \item Fang et al. \cite{fang2019diagnosis} developed an information theory approach informed by a KG to select features for training machine learning models. The goal of their study was to develop a predictive model of Chronic Obstructive Pulmonary Disease (COPD) from a tabular dataset including twenty-eight features representing medical tests and patient symptoms. First, a KG was constructed by integrating electronic medical records (EMRs) and domain-specific biomedical knowledge to identify and represent relationships among diseases, symptoms, causes, risk factors, drugs, side effects, and more. The features of the tabular dataset corresponded to nodes in the KG. Their algorithm, CMFS-$\eta$, uses the weights between features in the KG to iteratively add or remove features from the set according to an information-theory-based heuristic. The study used this approach to select subsets of the corresponding features of the tabular dataset to train an SVM model. 
    \item Ma et al.  \cite{ma2020knowledge} sought to develop a model to predict whether a given Android app contained malware based on the Android API calls contained in the source code. First, they used the official documentation to construct a KG containing all API entities, such as classes and methods, as well as relationships between entities, such as return types and inheritance. Next, they identified a set of permissions considered to be highly sensitive that was required for each API entity. The study created a binary feature vector for each application based on whether or not a given entity was present in the code. To reduce the size of the binary feature vector, the authors selected only entities that were between one to four hops from a node requiring sensitive permission. <As not all entities contained explicit links in the documentation, an LSTM model was used to identify an additional subset of entities that shared similar descriptions with entities that require sensitive permissions.> This feature vector could then be used to train a classification model. A detailed description of how sensitive APIs, nodes in the KGs, are selected is shown in Figure~\ref{fig:andriod}.

    \begin{figure}[!ht]
    \centering
    \includegraphics[scale=0.12]{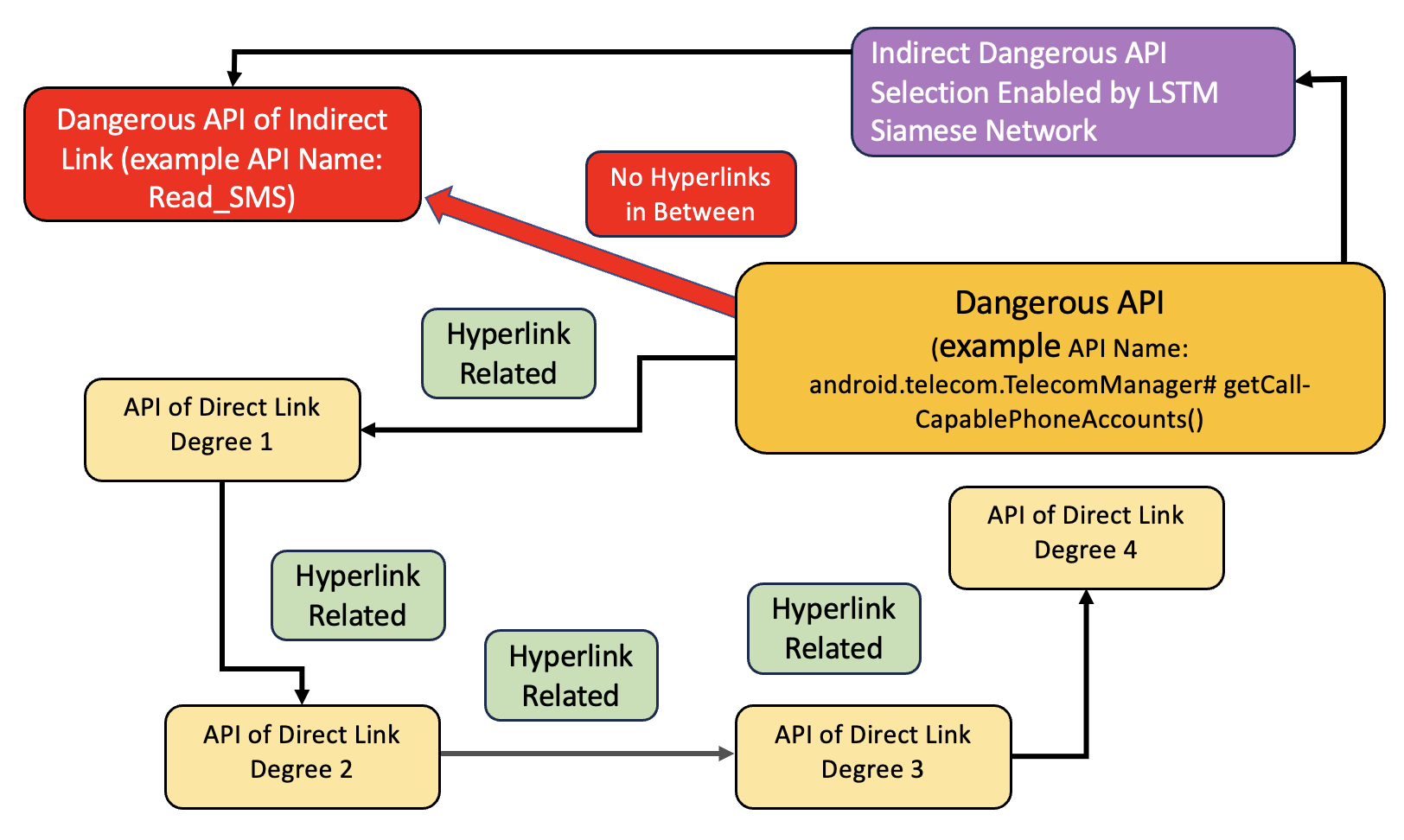}
    \caption{Example of Direct and Indirect Dangerous APIs Selection Enabled by the Android API KG. The golden-orange rounded rectangle in the figure signifies a dangerous API called ``getCall-CapablePhoneAccounts," which facilitates the retrieval of Phone Account Handles for making and receiving calls. The light-yellow rounded rectangles are APIs directly connected to the Dangerous API, up to four degrees of separation through hyperlinks, with the understanding that links beyond this do not markedly enhance classification accuracy. The Siamese-BiLSTM network comes into play by identifying indirectly connected, potentially dangerous APIs—represented by the red rounded rectangle, such as ``READ SMS," which allows reading SMS messages but lacks a direct hyperlink or descriptive connection to other APIs. By embedding API descriptions into a vector space using Word2Vec and processing them through a Bidirectional LSTM, the network encodes the APIs' textual data from both directions for a full context capture. These encoded vectors are then condensed through a dense layer into a final representation. Comparing these representations enables the network to detect hidden APIs that, while not directly linked, share sensitive characteristics with the known dangerous API, thereby revealing hidden dangers through textual similarity rather than explicit interlinking.}
    \label{fig:andriod}
\end{figure}

\item Jaworsky et al. \cite{jaworsky2023interrelated}, developed an unsupervised approach for selecting significantly interrelated features and eliminating redundant features from a KG, which they applied to a health survey dataset published by the Behavioral Risk Factor Surveillance System (BRFSS) \citep{centers2014behavioral}. The algorithm works by iteratively ranking scores based on their connections. This feature selection approach is divided into four main steps outlined in Figure~\ref{fig:health_survey}. 

    \begin{figure}[!ht]
       \centering
         \includegraphics[scale=0.125]{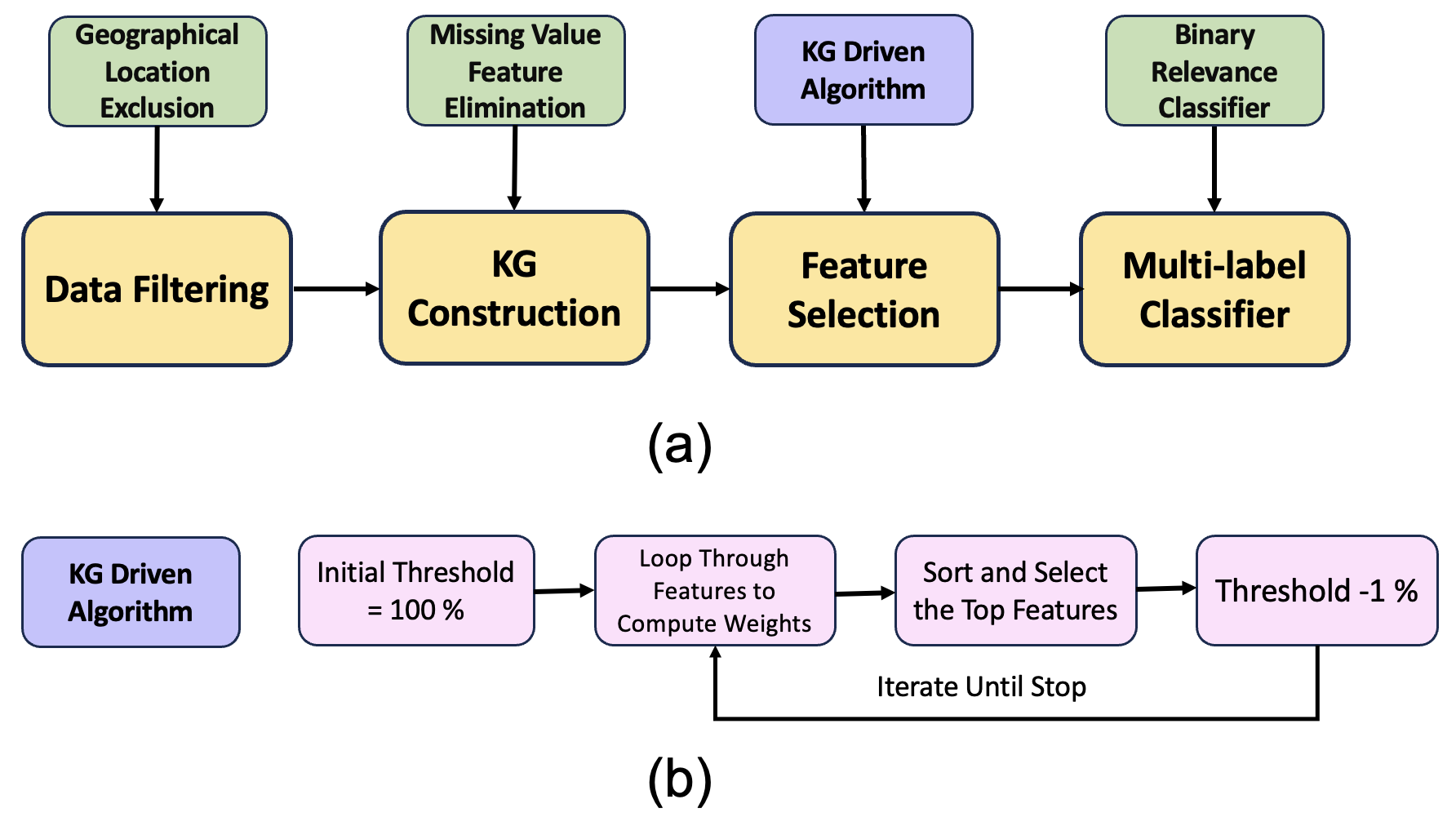}
        \caption{Illustration of interrelated feature selection procedure. 1) In the data filtering step as shown in part (a), states lacking lung cancer cases are excluded after referencing previous surveys spanning several years. 2) Features with over 50\% missing values are eliminated. Then a KG is constructed from the remaining features. 3) A KG driven algorithm is used to transform the health survey question list to a data set with significantly interrelated features. 4) Finally, a binary relevance classifier (a special case of multi-label classifier) is proposed to predict the likelihood of multiple diseases by identifying 1-to-many cancer relationship. In part (b), the KG driven algorithm starts with the initial threshold 100\%. Then it loops through existing features and compute weights for each (features with more edges will get more weights). By sorting the weights, the features with highest weights are kept and the threshold is subtracted by 1\%. The algorithm is iterated until the stopping criterion is met.}
        \label{fig:health_survey}
     \end{figure}
 \item In the Hadith Corpus KG created by Mohammed et al. \cite{atef2022feature}, nodes represent distinct features and semantic categories derived from Hadith texts. Features include specific Islamic terms like ``prayer" or ``fasting," while categories encompass broader thematic areas like rituals, ethics, jurisprudence, and other domains of Islamic scholarship. Edges in this KG quantify associations between features and categories based on co-occurrence frequency.

 \begin{figure}
     \centering
     \includegraphics[scale=0.25]{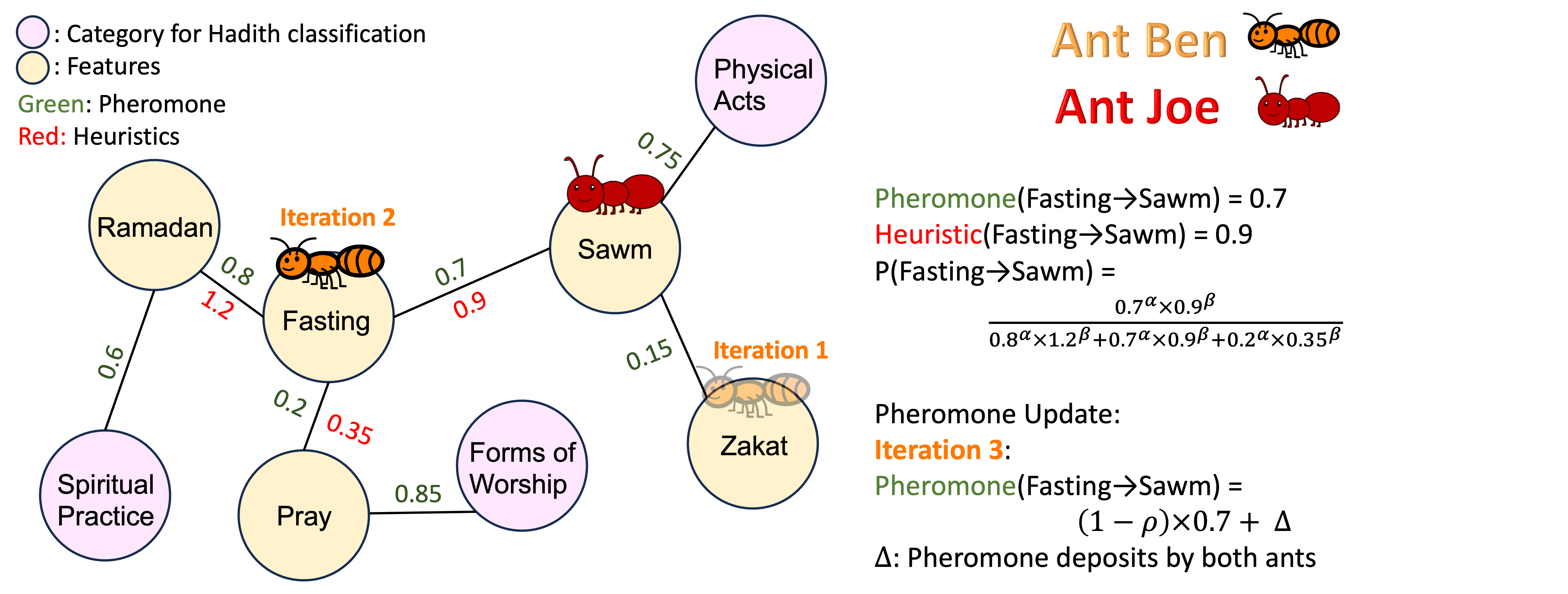}
     \caption{A demonstration of simplified ACO (Ant Colony Optimization) feature selection on Hadith Corpus KG. Here, two ants named Ben and Joe traverse the KG, with Ben starting at the ``Zakat" node and moving to ``Fasting"
     across iterations, and Joe beginning his journey at a randomly selected node ``Sawm". The pheromone and heuristic values, represented by the green and red numbers above and below the edges, are aggregated outcomes of the explorations conducted by all ants in the system. Parameters $\alpha$ and $\beta$ determine the relative influence of pheromone trails and heuristic information, respectively, while the evaporation rate $\rho$ ensures flexibility in pathfinding, preventing premature convergence on suboptimal routes. The collective pheromone deposit $\Delta$ between ``Fasting" and ``Sawm" by Ben and Joe is a cumulative measure reflecting the alignment of the Hadith content with specific categories, denoted by the pink nodes. The probability that Ben chooses ``Sawm" as the next feature is computed as a normalized version of $\text{Pheromone}^\alpha\times\text{Heuristics}^\beta$ (see the middle right of the figure). In this instance, the focus is to reinforce the linkage between fasting-related Hadiths and the ``Physical Acts" category, differentiating it from the ``Spiritual Practice" category and the ``Forms of Worship" category, which are more aligned with spiritual benefits and devotional acts. }
     \label{fig:ACO}
 \end{figure}
 
 Feature selection for text classification is guided by Ant Colony Optimization (ACO) \citep{dorigo2005ant,dorigo2006ant,dorigo2019ant}. ACO is a probabilistic technique for solving computational problems which can be reduced to finding good paths through graphs. Inspired by the behavior of ants finding the shortest path from their colony to food sources, ACO is a part of swarm intelligence methods and a subset of evolutionary algorithms.  Initially, several paths are randomly constructed, and after traversing a path, an ant deposits Pheromones along it (typically inversely proportional to path length), so shorter paths receive more pheromones.  Over time, the pheromones evaporate, reducing their attractive strength to prevent premature convergence.  When choosing their paths, ants probabilistically prefer paths with stronger pheromone concentrations while also exploring new paths to avoid local optima. The process is repeated until convergence. In this way, ACO balances between exploring new feature paths (exploration) and intensifying the search around promising features found in previous iterations (exploitation), adapting dynamically to find optimal feature sets for text classification \citep{parpinelli2002data,martens2007classification,aghdam2009text,onan2023srl}.   The pheromone trail and PageRank-like heuristic measure guide this optimization.   We provide a graphical illustration of the ACO feature selection process in Figure~\ref{fig:ACO}.
 
 This study demonstrates that integrating ACO into Arabic text classification yields a notable 3\% average increase in accuracy, F1 score, recall, and precision compared to conventional methods like Naive Bayes, Random Forest, Decision Trees, and XGBoost, contributing significantly to the field of Arabic text classification.
\end{itemize}

\subsection{Data Linking and Data Integration-Similarity Based Methods}
\begin{figure}
    \centering
    \includegraphics[scale=0.125]{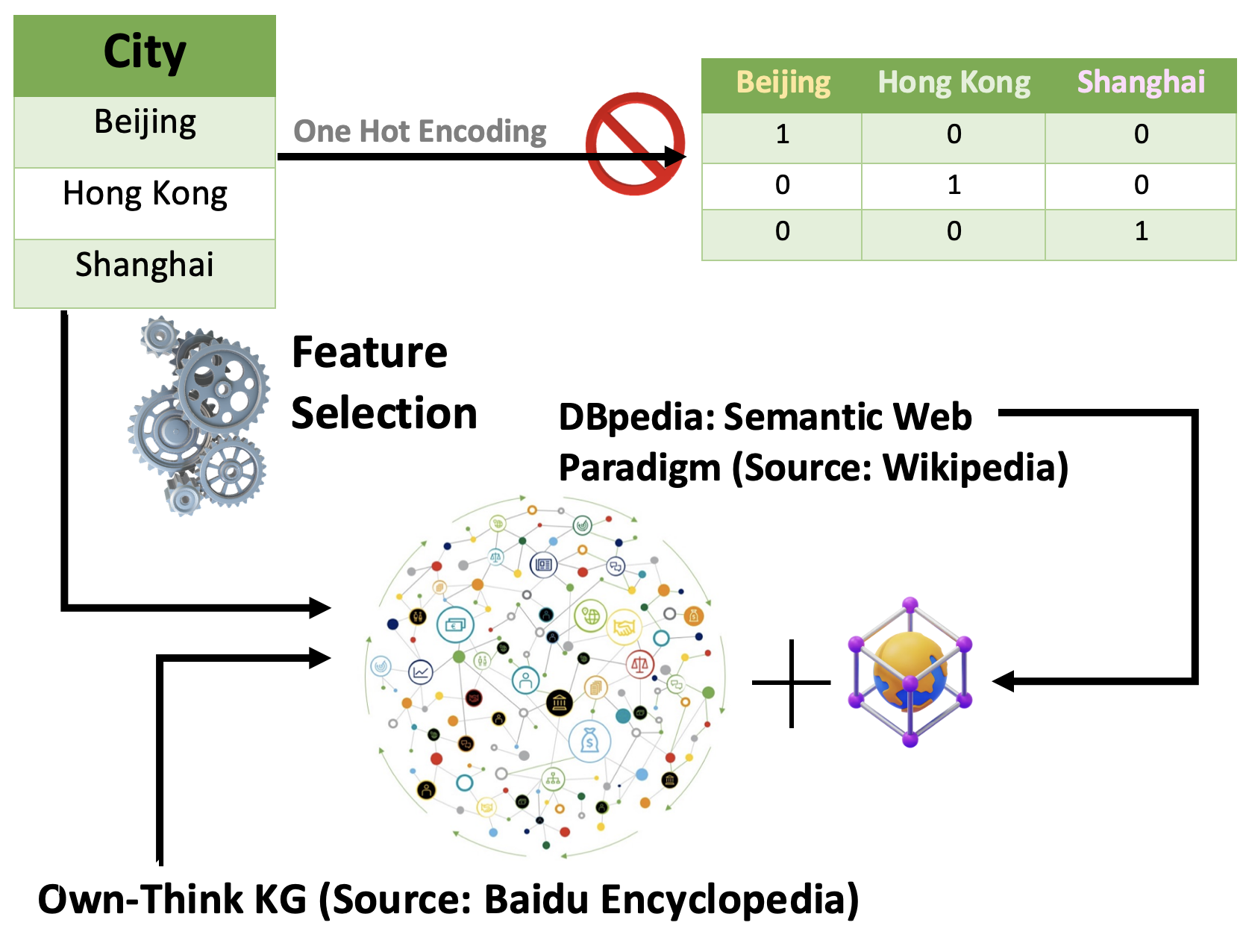}
    \caption{Own-Think KG Advantage over Tradition One-hot Encoding. Consider a dataset that includes information about various cities, Beijing, Shanghai, and Hong Kong, where each city is represented by non-numerical discrete features such as its name. In a traditional dataset, this name might be converted into a numerical form using techniques like one-hot encoding. However, this process strips the city's name of any contextual information about the city itself.  Using a KG like the Own-Think KG, we can query additional information about each city to enrich the features, such as geographical, economic, demographic, cultural features, and so on. }
    \label{fig:own_think}
\end{figure}

\begin{itemize}
\item Data linkage and data integration refer to the process of combining different sources of data\cite{chang2018making}. As KGs are developed to summarize large data, they can be great, easy-to-use tools for adding additional data and context to make ML workflows. For example, features of a given dataset can be expanded to include additional information per sample based what we know about a given feature. In Li et al. \cite{li2020feature}, the study collected data on self-reported student anxiety levels as well as basic information such as age, gender, grade, and home address. They then used the ``Own-Think KG" (see Figure~\ref{fig:own_think}) as well as ``DBpedia," both known for their credibility and encyclopedic nature, to identify other features for their analysis based on the home address, including weather, population size, and GDP at both the district and regional area levels. These KGs follow a clear and explainable three-tuple storage structure, consisting of entities, attributes, and values, making them suitable for non-numerical feature generation. Importantly, they offer online querying capabilities, eliminating the need to download extensive datasets \citep{auer2007dbpedia}. 

\end{itemize}

\subsection{Knowledge Graph Embeddings-Vector Embeddings}

The embedding-focused approach in feature selection, exemplified by methods like the DistMult \cite{yang2014embedding}, ComplEX \citep{trouillon2016complex}, TransE \citep{bordes2013translating}, and RESCAL \citep{nickel2011three}, and FeaBI \cite{ismaeil2023feabi}, RippleNet \citep{wang2018ripplenet} seeks to represent nodes in a continuous vector space that capture deep semantic relationships and properties. This is a similar concept to word embeddings. Whereas in word embeddings, similar vectors capture similar semantic meaning, with similar words having similar representation, graph node embeddings capture relationship similarity within the graph network. The approach is popular for various applications, including link prediction \citep{kumar2020link} and entity classification \citep{al2020named}. Link prediction serves several purposes, from selecting movies a user would be interested in, to predicting drug-target interactions. Several methods have been developed to leverage embeddings for recommendation algorithms.
\begin{figure*}[ht]
  \centering
  \begin{minipage}{0.32\textwidth}
    \includegraphics[width=\linewidth]{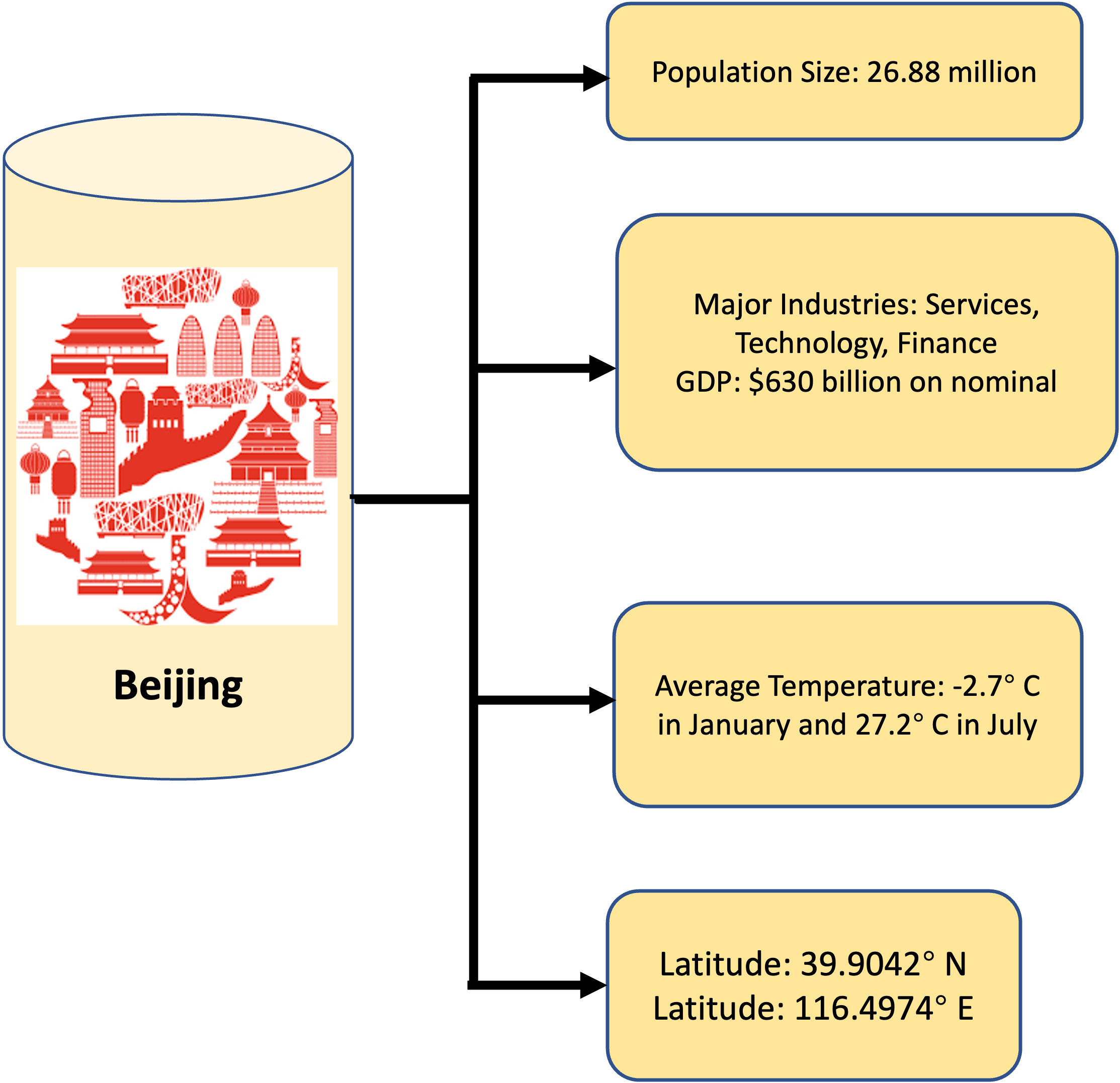}
    %\caption{First figure}
    %\label{fig:beijing}
  \end{minipage}\hfill
  \begin{minipage}{0.33\textwidth}
    \includegraphics[width=\linewidth]{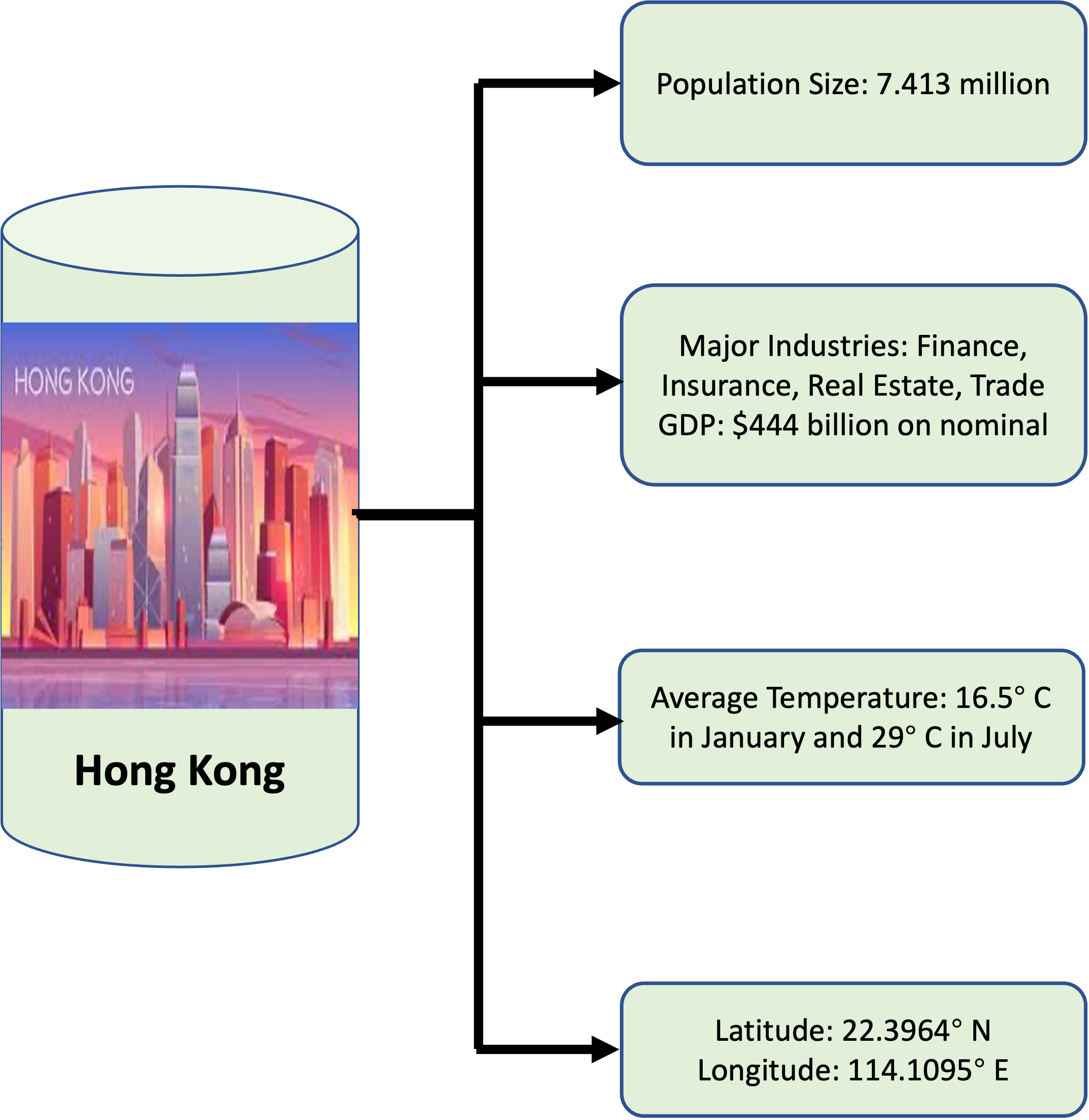}
    %\caption{Second figure}
    %\label{fig:figure2}
  \end{minipage}
    \begin{minipage}{0.32\textwidth}
    \includegraphics[width=\linewidth]{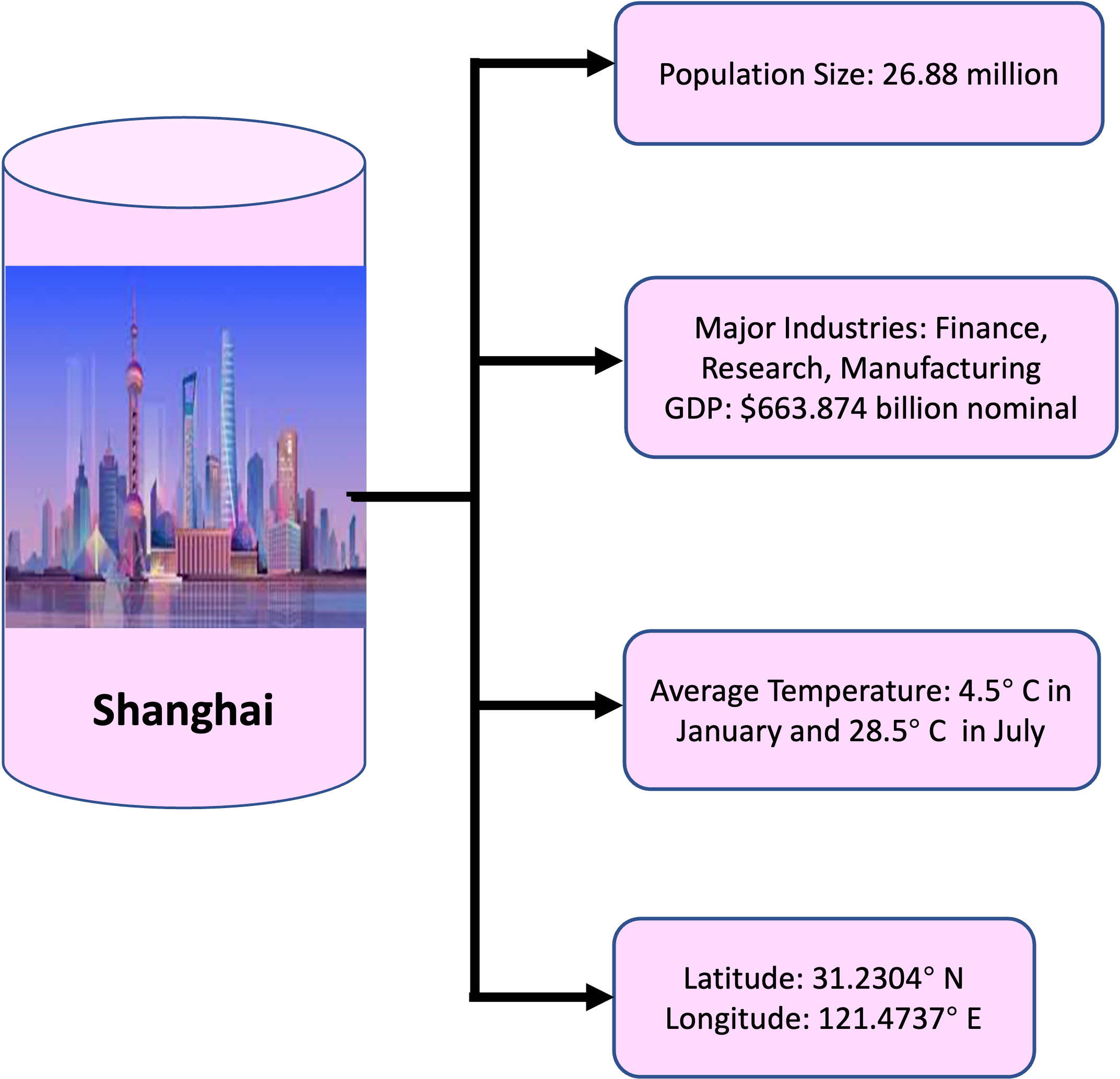}
    %\caption{Second figure}
    %\label{fig:shanghai}
  \end{minipage}
\caption{Demonstration of Non-numeric Discrete Features Enrichment and Selection by Own-think KG. The figure includes enriched information for Beijing, Hong Kong, and Shanghai. For example, the additional features for Shanghai provided by the Own-Think KG (see Figure~\ref{fig:own_think}) detail Shanghai's population size, average temperature, latitude, longitude, and GDP. contribute to a richer, more nuanced profile of Shanghai, compared to a one-hot encoding representation of each city, and offer additional insight as to how each aspect of a city may relate to the analysis at hand.}
    \label{fig:Own-thinkKGCities}
\end{figure*}
\textit{Embedding via DistMult:}
\begin{enumerate}
    
    \item  The DistMult method, designed to  predict missing relationships or facts within a KG \citep{chen2020knowledge}, embeds entities and their interactions as vectors, inherently performing feature selection by:
    \begin{itemize}
        \item \textit{Capturing Semantic Similarities:} Entities with closer interactional kinship within the KG are embedded proximately, emphasizing features underlying these semantic similarities.
        \item \textit{Highlighting Relevant Interactions:} DistMult accentuates features defining the interactions, such as biological pathways or chemical properties relevant to the interaction.
    \end{itemize}
    
    \item \textit{Optimization of Feature Representation:} The DistMult training process fine-tunes the entity and relation representations in the vector space, adjusting the significance of various attributes to enhance model accuracy.
\end{enumerate}

\begin{itemize}
\item One relatively simple strategy for edge prediction is to first create embeddings for each node and then train a classification algorithm to predict whether or not a connection exists between two nodes given their embeddings. For example, Wang et al. \cite{wang2022kg} utilized this strategy to predict drug-target interactions. In this study, the authors created node embeddings from a KG that contained known drug-target interactions. Next, they trained a deep learning model that took in a pair of embeddings (one drug and one target) to predict whether or not this pair was an existing edge in the graph. The authors showed that the model was able to identify some known interactions that were removed from the training set.

%for Ripp-MKR/RippleNet, I think it makes sense to start with RippleNet since they were the first to use embeddings in this way. ripp-mkr extended this
%the RippleNet paper cites several examples of embeddings for recommendation algorithms that we could also use.
% feels like another section on path based methods would have been helpful prior to ripplenet?

%Quan Wang, Zhendong Mao, Bin Wang, and Li Guo. 2017. Knowledge graph embedding: A survey of approaches and applications. IEEE Transactions on Knowledge and Data Engineering 29, 12 (2017), 2724–2743.

 % 

\item  A unique example comes from Wang et al., who proposed a hybrid KG embedding and path-based method in a recommendation algorithm they named RippleNet \citep{wang2018ripplenet}. In this context, the KG contains nodes representing items that can be recommended, for example, movies, along with other nodes representing other features associated with each item, such as actors, genres, or release date. Edges associations between items and features, for example, a movie and its actors. In addition, there is a separate matrix that contains the interactions between each user and item. The goal is to predict the likelihood of a user selecting an item given the KG and the user's prior interactions. The algorithm begins by initializing the representation of each item based on the user's click history. Next, the algorithm iterates over items that are increasing hops from items the user had already interacted with. The end result is an embedding of the relevance of each item that is combined with the initial vector representation with a model for the final prediction of the likelihood of selecting that item. This was later extended by Wang et al. \cite{wang2021multitask} by having a combined deep framework that is simultaneously trained on a KG embedding task in addition to learning the recommendation task. The model architecture features shared latent features between the two tasks, with the idea being that the inclusion of the embedding task will enhance the latent representations. We give an illustration of Ripp-MKR in Figure~\ref{fig:ripp-mkr}.

\begin{figure}
    \centering
    \includegraphics[scale=0.28]{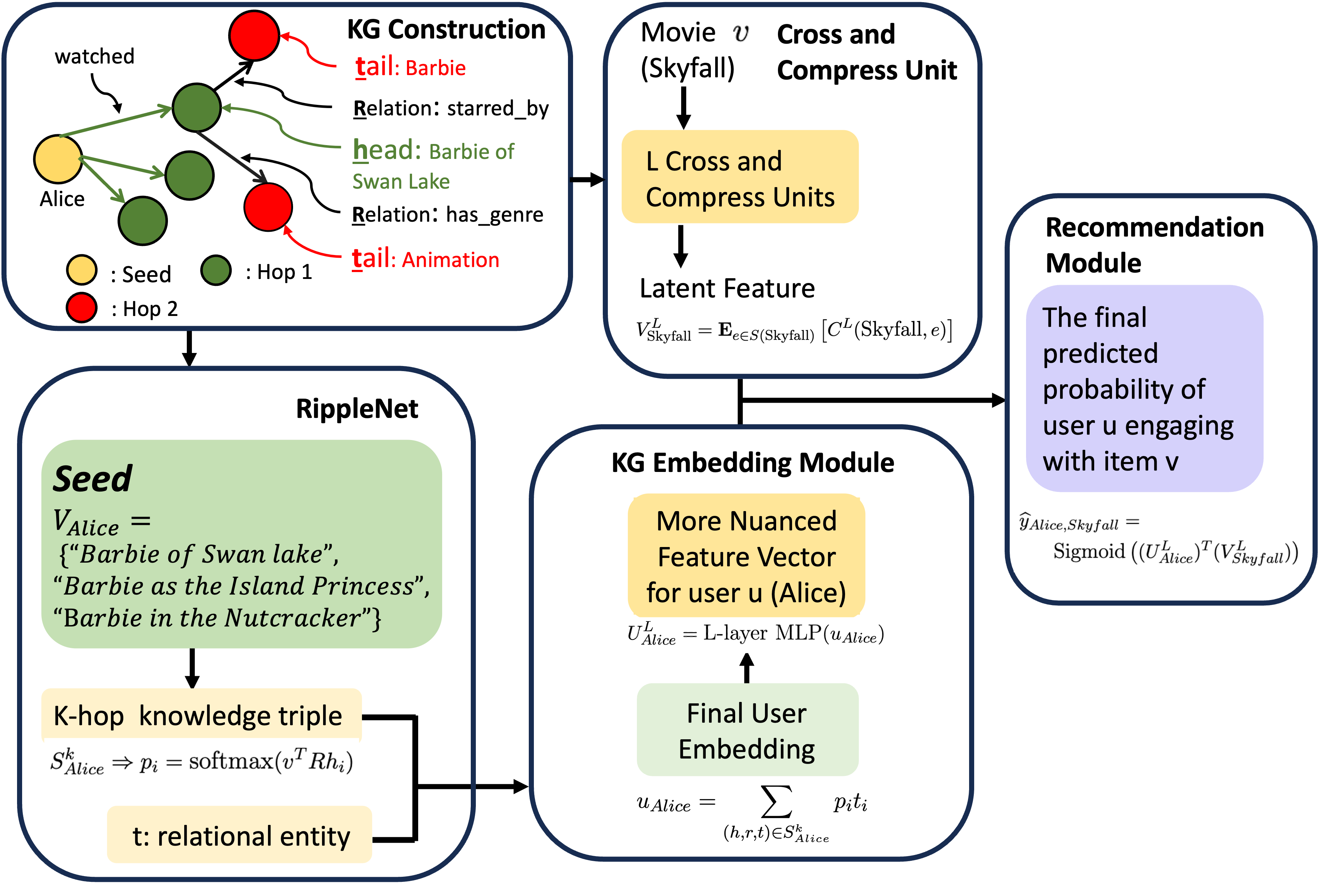}
    \caption{Illustration of Ripp-MKR Feature Learning Mechanisms. The Ripp-MKR model involves a recommendation system KG with nodes representing users, movies, genres, and actors. In this KG, relationships such as ``Alice watched Barbie of Swan Lake," ``Barbie of Swan Lake is starred by Barbie," and ``Barbie of Swan Lake has genre Animation" are examples of how the system is structured (see \textbf{KG Construction}). Taking Alice as the initial point, we construct a historical set, $V_{\text{Alice}}$, comprising Alice's movie-watching history, which includes three movies (see \textbf{RippleNet}). RippleNet then extends Alice’s preference for the Barbie series to other movies with similar genres and actors, like ``Barbie." The \textbf{KG Embedding Module} (KGE) refines Alice's embedding, $u_{\text{Alice}}$, by aggregating all the $k$-hop softmax-weighted tail embeddings $t_i$, for instance, ``Animation" and ``Barbie" (see \textbf{KG Construction}). This refined embedding, $u_{\text{Alice}}$, is processed through an $L$-layer MLP to derive a nuanced user vector, $U_{\text{Alice}}^{L}$. The KGE is informed by the interactions among movies, genres, and actors. The \textbf{Cross and Compress Unit} (CCU) examines the interactions between different genres by calculating the outer product of the movie vector $v$ (e.g., ``Skyfall") and an entity vector $e$ from the set $S_{\text{Skyfall}}$, which includes entities related to ``Skyfall" in the KG. After performing the outer product between $v$ and each $e \in S_{\text{Skyfall}}$ $L$ times, the final latent feature vector, $V_{\text{Skyfall}}^{L}$, for ``Skyfall" is obtained by taking the expectation over the L outer products. The \textbf{Recommendation Module} then selects the movie with the highest sigmoid probability from the inner product of $U_{\text{Alice}}^{L}$ and $V_{\text{Skyfall}}^{L}$, denoted by $\widehat{y}_{\text{Alice,Skyfall}}$. From potential next movies like ``Skyfall," ``Inception," and ``Barbie: Fairytopia," Ripp-MKR recommends ``Barbie: Fairytopia" to Alice as it has the highest probability value, indicating it as the most suitable next watch.}

    \label{fig:ripp-mkr}
\end{figure}

\item Ismaeil et al. \cite{ismaeil2023feabi} introduced a method, FeaBI, to generate interpretable KG entity embeddings. First, a standard KG embedding is calculated. Additionally, a few categories of features for each node are extracted to form a vector, including the types of edges or relations it has, the types of nodes it is connected to, sequences of edge types of a certain length, and graph structural statistics. Next, regression random forest models are trained to predict each of the original embedding dimensions from its extracted feature vector. The random forest model ranks features based on their importance for the reconstruction task. These rankings can be used to better understand the information captured by embeddings. Additionally, a smaller subset of the feature vector can be selected for the most important features and used in place of the original embedding for more interpretable analysis. 

% \cite{steenwinckel2022ink}

% and for explaining KG embedding models through selecting important KG features.

% \item Ismaeil et al. \cite{ismaeil2023feabi} introduced a method, FeaBI, to generate interpretable KG entity embeddings. It first extracts propositional features that are expressed as propositions about entities from KG and expresses them in Description Logic \citep{baader2004description}. These extracted features are then used to construct Boolean vectors for each entity based on its neighborhood in the KG. Four types of features are constructed, namely, relations, relations with entities \cite{steenwinckel2022ink}, graph structural statistics, and surrounding entities. Then, a regression random forest, trained using a pre-computed embedding model, with feature importance rank is applied to perform feature selection. The goal of this model is to reconstruct the embedding-based entity representations using the features defined in the first step. The model ranks features based on their importance for the reconstruction task. Features that receive the highest scores are considered as explanations for the KG embeddings.

\end{itemize}

% \subsection{Similarity-based methods}

% Similarity-based methods, expanding with other applications, involve comparing entities and attributes within a graph to identify similarities and differences using metrics like cosine similarity or Jaccard index. These methods excel in identifying closely related entities or attributes, beneficial for clustering or recommendation systems. 

% For instance, Ma et al. \cite{ma2020knowledge} combine KGs with NLP tools and Bi-LSTM Encoder with a Siamese Network to improve Android malware classification. Similarly, Jaworsky et al. \cite{jaworsky2023interrelated} applies a qualitative approach to feature selection in health survey datasets, leveraging KGs for identifying significant interrelated features and eliminating redundancy, enhancing the effectiveness of health advice through a binary relevance classifier. These examples illustrate the adaptability and potency of similarity-based methods in various domains, from technology to healthcare, by leveraging KGs for more nuanced and effective feature selection and classification.

\subsection{Deep Learning-Advanced Network representation Learning}

Deep Learning models are designed to capture high-level, abstract representations of data. This ability allows them to capture meaningful insights from KGs, thereby enhancing applications in various domains, including personalized recommendations and predictive healthcare analytics.

% The advanced network representation learning segment delves into the utilization of DL models to interpret and analyze knowledge within KGs through complex network structures, such as DDKG and KGFLEX. These models are specifically designed to capture high-level, abstract representations of data, facilitating sophisticated feature selection and unveiling intricate patterns and relationships. KGFLEX, for instance, employs multi-hop predicates and evaluates feature significance through entropy and information gain, optimizing recommendation systems. DDKG, on the other hand, focuses on drug-drug interaction predictions by integrating molecular data and employing attention mechanisms to refine drug embeddings. Both models exemplify the advanced capabilities of network representation learning in extracting meaningful insights from KGs, thereby enhancing applications in various domains, including personalized recommendations and predictive healthcare analytics. Finally, the powerful GNN framework provides another insight to strength the users' toolbox for feature selection on KGs.

\begin{itemize}
    \item Anelli et al. \cite{anelli2021sparse} proposes KGFlex, a recommendation system \citep{shani2011evaluating} that integrates KG-based feature selection to improve the personalization and accuracy of recommendations. They use the notion of multi-hop predicates \citep{zhang2021cone} (i.e., considering chains of predicates that connect two entities at a high depth) to construct the semantic features on a KG. For instance, $\text{A}\rightarrow \text{B}\rightarrow \text{C}$ is a 2-hop predicate. 
    
    In the feature selection step,  KGFlex utilizes the concepts of entropy and information gain \citep{shannon1948mathematical, rokach2005top} to assess how significant and relevant a feature is to a user when determining whether to engage with an item or not, i.e., to watch a movie or not. The features, represented as $\langle$predicate$,  $entity$\rangle$ pairs, are then embedded in a latent space to construct the user-item interaction along with user embeddings via DL methods. For a particular user, the items with higher user-item interactions are recommended. All the embeddings and model parameters in KGFlex are learned from the Bayesian Personalized Ranking (BPR) optimization criterion \citep{rendle2012bpr}. The whole procedure is visualized in Figure~\ref{fig:sparse}.

    \begin{figure}
        \centering
        \includegraphics[scale=0.32]{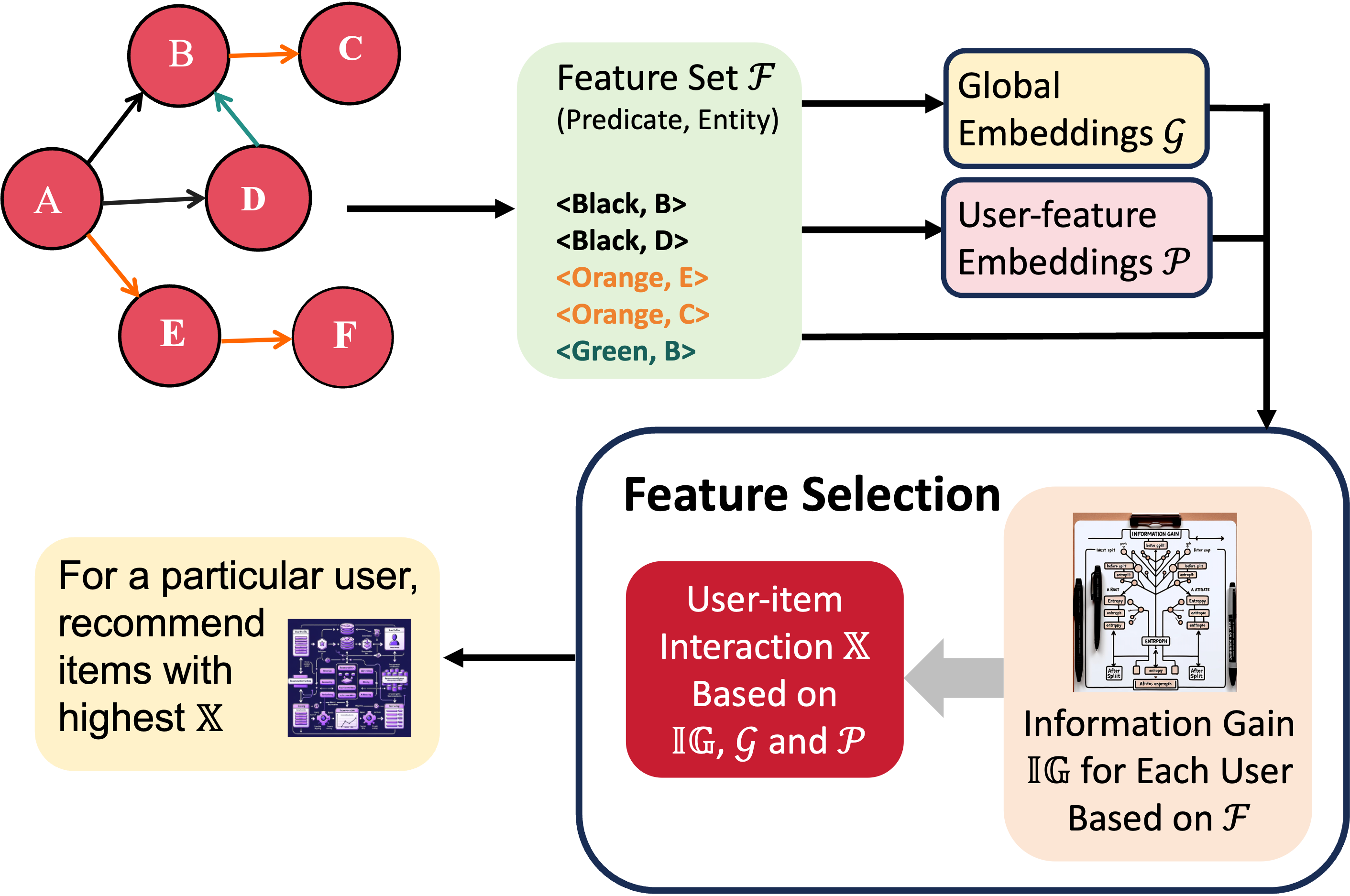}
        \caption{Illustration of KGFlex feature selection and recommendation procedure.  We start with a KG with 6 nodes and 6 predicates (edges/relations). A feature set $\mathcal{F}$ is constructed where each element is of form $\langle\text{predicate, node}\rangle$; for instance, from node A we can get to node B via a black predicate, then a feature is constructed as $\langle$Black, B$\rangle$. We construct a global embedding set $\mathcal{G}$ representing each feature in $\mathcal{F}$, and a user-feature embedding set $\mathcal{P}$ for each pair of user and feature. All embeddings and parameters in KGFlex are learned via DL methods with the BPR optimization criterion.  We then associate each user-feature pair with an information gain $\mathbb{IG}$,  which measures the expected reduction in information entropy from a prior node to a new node that acquires some information. For instance, suppose a user is currently at node $A$. The computed information gain $\mathbb{IG}$($\langle$Black, B$\rangle$)=1, $\mathbb{IG}$($\langle$Orange, E$\rangle$)=0 and $\mathbb{IG}$($\langle$Black, D$\rangle$)=1 means the nodes $B,D$ and the predicate ``Black" have influential impacts on the user's next move. Finally, for each user, we compute the user-item interaction $\mathbb{X}$ and recommend items to him with the highest $\mathbb{X}$ values. }
        \label{fig:sparse}
    \end{figure}

    The performance of KGFlex is evaluated on three datasets from various domains, \emph{Yahoo! Movies}, \emph{MovieLens}, and \emph{Facebook Books}. The experiments are designed to test the performance of KGFlex in terms of the Gini Index \citep{gastwirth1972estimation, castells2021novelty}). KGFlex outperforms certain latent factor models such as kaHFM \citep{anelli2019make}, Item-kNN \citep{koren2010factor}, NeuMF \citep{he2017neural} and BPR-MF \citep{rendle2012bpr} by an average of 18\%. It also surpasses other key metrics, such as Item Coverage \citep{adomavicius2011improving}, in the recommendations it generates. Additionally, it excels in metrics like ACLT \citep{abdollahpouri2019managing}, PopREO, and PopRSP \citep{zhu2020measuring}, which measure recommendation performance concerning the underrepresentation of rare items. It is occasionally outperformed only by kaHFM in top-10 recommendations. % It also outperforms in other key metrics like Item Coverage \citep{adomavicius2011improving} of the recommendations it generates as well as the its behavior on the underrepresentation of rare items (ACLT \citep{abdollahpouri2019managing}, PopREO and PopRSP \citep{zhu2020measuring}), occasionally being outperformed only by kaHFM  in top-10 recommendations.

    \item Su et al. \cite{su2022attention} presents an attention-based KG representation learning framework, named DDKG, aimed at feature representation and selection to improve drug-drug interaction (DDI) prediction. This approach allows for end-to-end prediction of DDIs. We summarize the DDKG into the below four main parts:
    \begin{itemize}
        \item[a.] \emph{KG Construction:} The KG construction amalgamates the Simplified Molecular Input Line Entry System (SMILES),  SMILES-associated triple facts, and entities such as proteins and diseases. For example, we have two drugs, A and B, and we integrate their SMILES sequences alongside their relationships (e.g., ``targets”) with diseases into the KG.
        \item[b.] \emph{Drug Embedding Initialization:} DDKG uses an encoder-decoder layer to learn the initial embeddings of drug nodes, mainly from the SMILES sequences in the KG. This step transforms the SMILES sequences of drugs A and B into vector representations that capture their chemical structure and properties.
        \item[c.]  \emph{Drug Representation Learning:} This part, consisting of three elements, serves as the key feature selection step in DDKG. %(See Figure 2 in \cite{su2022attention} for details).
          \begin{itemize}
            \item[-] \textit{Neighborhood Sampling:} For each drug node, a fixed-size set of neighboring nodes is selected. The significance of each neighbor is determined by \emph{attention weights}, which are calculated based on the embeddings of the nodes and the types of relationships among them. This step ensures only the most relevant neighbors (in terms of both graph structure and drug relationships) are considered for further computation.
            \item[-] \textit{Information Propagation: } It involves calculating a weighted sum of the neighbor embeddings. The attention weights (calculated in the previous step) are used to determine how much each neighbor’s information should contribute to the drug node’s new representation. This ensures that more relevant neighbors have a bigger impact on the final representation.
            \item[-] \textit{Information Aggregation:} The weighted sum of the neighbor embeddings is combined with the drug node’s initial embedding. A final global representation of a drug node is obtained. %
        \end{itemize}

        \item[d.] \emph{DDI Prediction:} For a queried pair of drugs, DDKG estimates their interaction probability by simply multiplying their final respective representations derived in \emph{c}.
    \end{itemize}

    \item In the work by Hsieh et al. \cite{hsieh2021drug}, a Graph Neural Networks (GNN) \citep{zhou2020graph} is employed to advance the feature selection (drug selection) process for COVID-19 treatment from a drug-target interaction network (see Figure~\ref{fig:gnn}). The authors first constructed a COVID-19 KG (see the top-left region in Figure~\ref{fig:gnn}) and generated embeddings using a GNN. The method involves transferring knowledge from another drug repurposing KG (see top-right region) and learning high-dimensional embeddings for drugs that encapsulate the complex pharmacological characteristics of drugs (see middle region). By utilizing a ranking model informed by Bayesian pairwise ranking loss, this approach prioritizes potential drug candidates for downstream tasks such as gene set enrichment analysis (see middle-left region), Retrospective in vitro drug screening (see middle-right region), etc. Top 22 promising drugs including  Aspirin, Acetaminophen and Teicoplanin are highlighted in the paper, demonstrating the rapid identification of candidate drugs for COVID‐19 treatment.

    \begin{figure}
        \centering
        \includegraphics[scale=0.275]{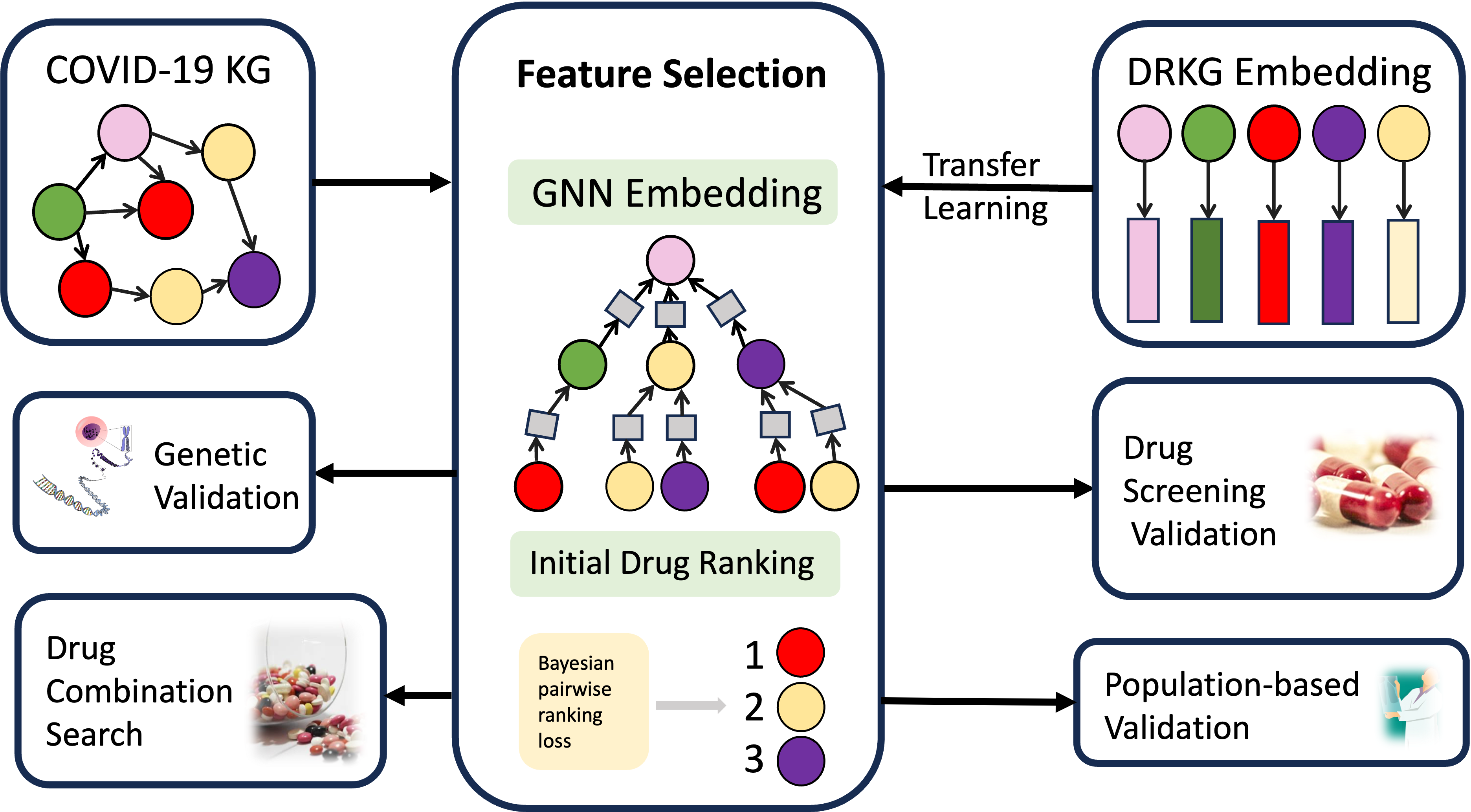}
        \caption{Feature selection (drug selection) via GNN embedding and drug ranking. The authors first construct a \textbf{COVID-19 KG} containing different types of nodes (including 3635 drugs) and interactions. The variational graph autoencoder with GraphSAGE messages passing \citep{kipf2016variational, hamilton2017inductive}, a specific type of GNN, was used to derive the drug embedding (the grey squares in \textbf{Feature Selection}) by transferring a drug repurposing KG (DRKG) \citep{zeng2020repurpose} to boost the representativeness. %(different colors in the \textbf{DRGK Embedding} box represent different embeddings for transfer learning)
        Initial drug ranking using Bayesian pairwise ranking loss is applied to rank and select possibly potent drugs out of all candidates, hence serving as a feature selection step. The model efficacy was demonstrated using different validations. For instance, the authors perform \textbf{Genetic Validation} by identifying significant associations between SARS-CoV-2 and selected drugs. \textbf{Drug Screening Validation} is also performed by retrospectively comparing selected drugs with effective drugs in various in vitro drug screening experiments. In the \textbf{Population-based Validation}, the proposed method identified six drugs administered to the COVID-19 patients out of ten positive drugs that were effective in the electronic health records. In addition, \textbf{Drug Combination Search} for improving the COVID-19 treatment efficacy is conducted on the selected drugs. All validation results testify the capability of the proposed method speeding up the discovery of candidate drugs for treating COVID-19.}
        \label{fig:gnn}
    \end{figure}
     
\end{itemize}

\subsection{Comparative Analysis of Different Approaches}
We evaluate the methodologies from referenced manuscripts, focusing on their advantages and disadvantages.

\textbf{1. Search Algorithms}
Used in the Hadith Corpus KG \cite{mohammed2020edge} with the ACO algorithm and in COPD diagnosis \cite{fang2019diagnosis} with the CMFS-$\eta$ algorithm. These methods highlight the importance of selecting appropriate strategies based on specific dataset requirements.

\textbf{2. Vector Embeddings} 
This approach, exemplified by the DistMult Algorithm and FeaBI, moves away from explicit path searches to embedding entities in a continuous vector space. It captures deep semantic relationships, facilitating the identification of intricate patterns relevant to complex domains like drug discovery \cite{dorigo2019ant, atef2022feature}.

\textbf{3. Similarity-based Methods}
These methods compare entities within a graph to identify similarities using metrics like cosine similarity or Jaccard index. They are beneficial for clustering or recommendation systems, as demonstrated by Ma et al. \cite{ma2020knowledge} in Android malware classification and Jaworsky et al. \cite{jaworsky2023interrelated} in health survey datasets.

\textbf{4. Advanced Network Representation Learning}
Utilizes deep learning models to interpret and analyze KGs, capturing high-level data representations. Examples include KGFLEX for optimizing recommendation systems and DDKG for drug-drug interaction predictions, showcasing the power of GNN frameworks in feature selection \cite{fang2019diagnosis}.

\textbf{Comparison and Contrast}
Search algorithms and similarity-based methods provide direct, interpretable insights into KG structures, making them suitable for applications requiring clarity and precision. In contrast, vector embeddings and advanced network representation learning offer a nuanced understanding of data, identifying complex patterns and relationships. These latter methods are valuable for scenarios where data relationships are not straightforward, enabling flexible and powerful KG modeling for predictive analytics. The drug ranking technique by Hsieh et al. \cite{hsieh2021drug} demonstrates the intersection of vector embeddings and advanced network learning, highlighting their transformative potential in feature selection.

\begin{table*}[t]
\caption{Comparison of Feature Selection Methods for KGs}
\centering
\begin{tabularx}{\textwidth}{@{} >{\hsize=.5\hsize}X >{\hsize=1.1\hsize}X >{\hsize=1.1\hsize}X @{}}
\toprule
\textbf{Method} & \textbf{Pros} & \textbf{Cons} \\
\midrule
\textbf{Search Algorithms} & 
Efficient and precise in known domains. Straightforward implementation. & 
May miss novel connections. Less adaptive to new patterns. \\
\addlinespace
\textbf{Vector Embeddings} & 
Captures deep semantic relationships. Scalable to large KGs. Enhances predictive power. & 
Challenges in interpretability. High initial training cost. \\
\addlinespace
\textbf{Similarity-based Methods} & 
Easy to understand. Efficient for clustering/recommendations. & 
Reliant on similarity metric quality. Computational challenges with large KGs. \\
\addlinespace
\textbf{Advanced Network Representation Learning} & 
Learns complex representations. Integrates heterogeneous data. Versatile in application. & 
Computationally intensive. Complex model structure. \\
\bottomrule
\end{tabularx}
\label{table:Comparison}
\end{table*}

\section{Challenges and Opportunities in Knowledge Graph Feature Selection}
\label{sec:challenges_opportunities}

Knowledge Graphs (KGs) are transforming data-driven fields like biomedical research, bioinformatics, and recommendation systems. They offer significant analytical capabilities but also present challenges and opportunities, especially in feature selection for machine learning models.

\subsection{Challenges}

Feature selection in KGs faces several hurdles:

\begin{enumerate}
    \item \textbf{High Dimensionality and Complexity:} KGs encompass numerous entities and relationships, creating high-dimensional spaces that challenge traditional feature selection methods.
    
    \item \textbf{Data Heterogeneity:} KGs integrate diverse data types (numerical, categorical, textual) from various sources, necessitating robust feature selection techniques.
    
    \item \textbf{Interpretability:} Enhancing interpretability is crucial, especially in fields like healthcare, where understanding why features are selected is essential.
\end{enumerate}

\subsection{Future Directions}
\label{sec:future_directions}

Several promising research avenues could redefine KG feature selection:

\begin{itemize}
    \item \textbf{Causal Inference Techniques:} Applying causal inference techniques to KGs can refine feature selection strategies \citep{malec2023causal}.
    
    \item \textbf{Embedding KGs into Feature Matrices:} Creating feature matrices from KGs facilitates downstream tasks and enhances model performance \citep{strande2017evaluating}.
    
    \item \textbf{Novel Algorithms:} Exploring algorithms like Ant Colony Optimization (ACO) introduces new approaches to feature selection within KGs \citep{dorigo2019ant, atef2022feature}.
    
    \item \textbf{Multi-objective Optimization:} Using multi-objective optimization techniques offers a refined methodology for feature selection, balancing criteria like redundancy and relevance \citep{mouret2015illuminating}.
    
    \item \textbf{Interdisciplinary Integration:} Combining KGs with quantum computing, reinforcement learning (RL), and federated learning (FL) can enhance feature selection. Quantum-enhanced selection addresses scalability, RL refines the process based on feedback, and FL enables decentralized selection, preserving privacy \cite{ma2021quantum, huang2022fedcke}.
    
    \item \textbf{Semantic Enrichment and XAI:} Leveraging the semantic information in KGs and applying Explainable AI principles can improve feature selection and model interpretability.
    
    \item \textbf{Domain Knowledge Integration:} Integrating domain-specific knowledge into the feature selection process results in more effective selections, particularly in specialized fields like genomics and pharmacology.
    
    \item \textbf{Multi-modal Data Fusion:} Combining various data sources into KGs offers a holistic view and unlocks new insights and applications.
    
    \item \textbf{Dynamic KGs and Real-time Feature Selection:} Developing methods for real-time feature selection as KGs evolve can lead to more agile models, critical in rapidly changing domains like social media analysis.
    
    \item \textbf{Collaborative KG Frameworks:} Creating frameworks for sharing and integrating KGs can enhance feature diversity and quality, fostering standardized protocols and benchmarks.
    
    \item \textbf{Ethical Considerations:} Prioritizing ethical considerations and bias mitigation in KG feature selection ensures fairness and equity in applications.
\end{itemize}

\section{Conclusion}
\label{sec:conclusion}

Examining these methodologies underscores the importance of scalability, accuracy, and interpretability in feature selection processes. As KGs grow, developing scalable algorithms that efficiently process large-scale KGs without losing information granularity is paramount. This requires a balanced approach that leverages KGs' rich semantic relationships while addressing computational challenges.

\section*{Key Points of the Paper}
\begin{itemize}
    \item Emphasizes combining feature selection techniques with KGs to enhance predictive modeling in biomedical research.
    \item Shows significant applications in bioinformatics, improving disease prediction and drug discovery processes.
    \item Discusses challenges like computational complexity and the need for comprehensive KGs, proposing future research to develop efficient algorithms and integrate additional data sources.
\end{itemize}

\section{Funding}
This work was funded by the National Institutes of Health (NIH) [U01 AG066833].
%USE THE BELOW OPTIONS IN CASE YOU NEED AUTHOR YEAR FORMAT.
%\newpage
%\appendix
\section*{Appendix}
\subsection{Appendix A. Table of Acronyms}
Table 2 lists the Table of Acronyms for this paper.
\label{appendix:acronyms}
\begin{table}[]
    \centering
    \begin{tabular}{l|p{5.5cm}}
    \hline\hline
    Abbreviation & Definition \\
    \hline
        ACLT & Average Coverage of Long Tail items \\
    ACO & Ant Colony Optimization \\
    AD & Alzheimer's Disease \\
    ADKG & Alzheimer's Disease Knowledge Graph \\
    AI & Artificial Intelligence\\
    AlzKb & Alzheimer's Disease Knowledge Base \\
    APOE & Apolipoprotein E \\
 AUC & Area Under the Curve \\
    Bi-LSTM & Bidirectional Long Short-Term Memory\\
    BPR & Bayesian Personalized Ranking \\
    COPD & Chronic Obstructive Pulmonary Disease \\
    CYP2D6 & Cytochrome P450 2D6 \\
    DDI & drug-drug interaction \\
    DistMult & The Distributed Multinomial Method \\
    DL & Deep Learning \\
    DR & Dimensionality/Dimension Reduction\\
    DSA-SVM & Direct Search Simulated Annealing with Support Vector Machine \\
    DTP & Drug-target Pairs \\
    %FS & Feature Selection \\
    GDB & Graph Database\\
    GNN & Graph Neural Network \\
    HMOX1 & Heme Oxygenase 1 \\
    KEGG & Kyoto Encyclopedia of Genes and Genomes \\
    KG & Knowledge Graph \\
    LDA & Linear Discriminant Analysis \\
    LLE & Local Linear Embedding\\
    ML & Machine Learning \\
    MLP & Multiple Layer Perceptron \\
    MQL & Metaweb Query Language \\
    MTHFR & Methylenetetrahydrofolate Reductase \\
    RDF & Resource Description Framework \\
    RFE & Recursive Feature Elimination \\
     nDCG &Normalized Discount Cumulative Gain \\
    NLP & Natural Language Processing \\
    NOS3 & Nitric Oxide Synthase 3 \\
    OWL & The Web Ontology Language \\
    PCA & Principal Component Analysis \\
    PPARG & Peroxisome Proliferator-Activated Receptor Gamma \\
    RDF & Resource Description Framework \\
    RFE & Recursive Feature Elimination \\
    RO & Relation Ontology \\
    TPI1 & Triosephosphate Isomerase 1 \\
    URIs & Uniform Resource Identifiers \\
    UMLS & Unified Medical Language System \\
    W3C & World Wide Web Consortium \\
    YAGO & Yet Another Great Ontology \\
    \hline\hline
    \end{tabular}
    \caption{Table of Acronyms}
    \label{tab:acronym}
\end{table}

\subsection{Appendix B. A more detailed description of KGs of sizes tiny, small, and medium}
\begin{figure}
    \centering
    \includegraphics[scale=0.1]{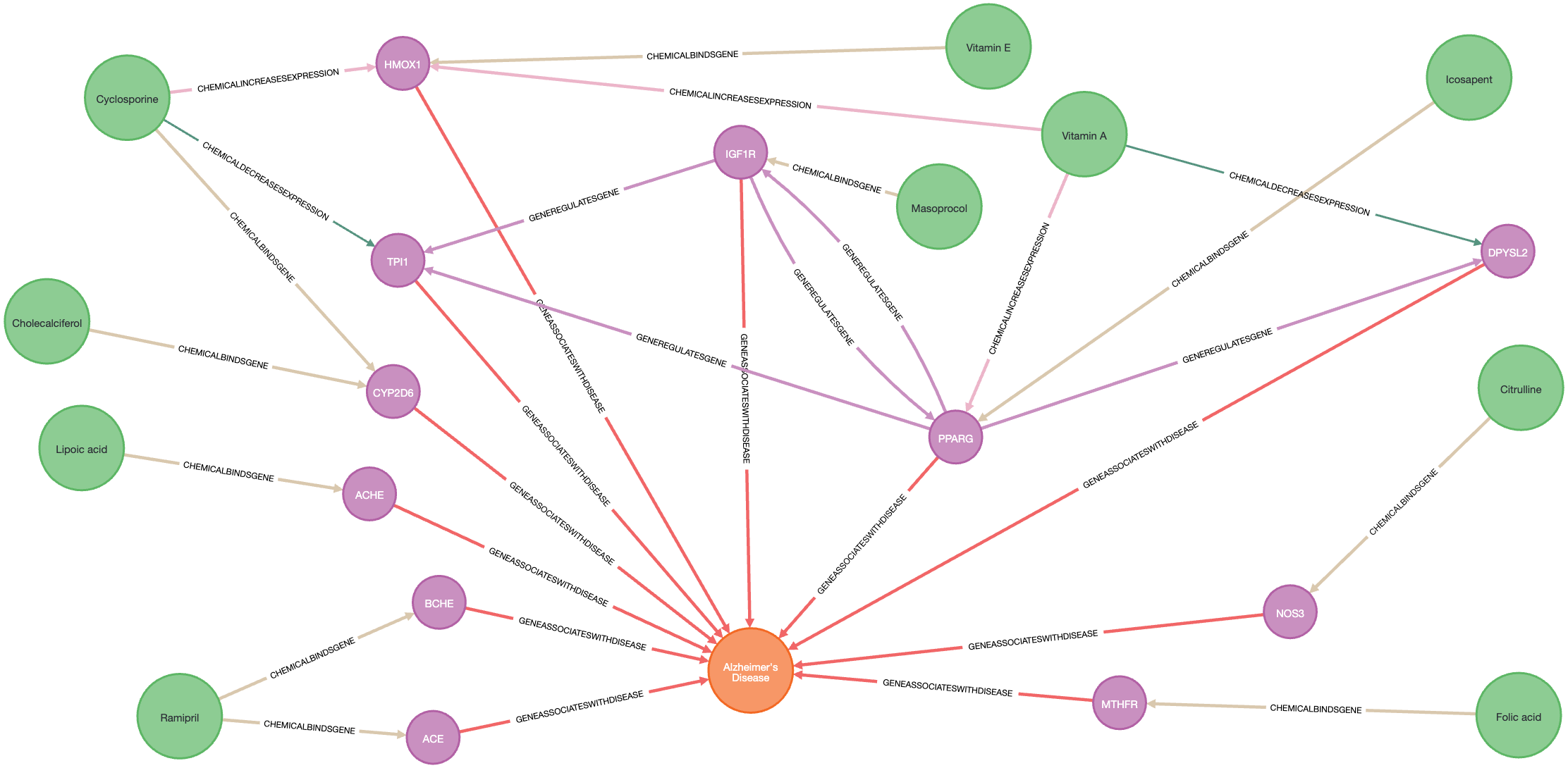}
    \includegraphics[scale=0.2]{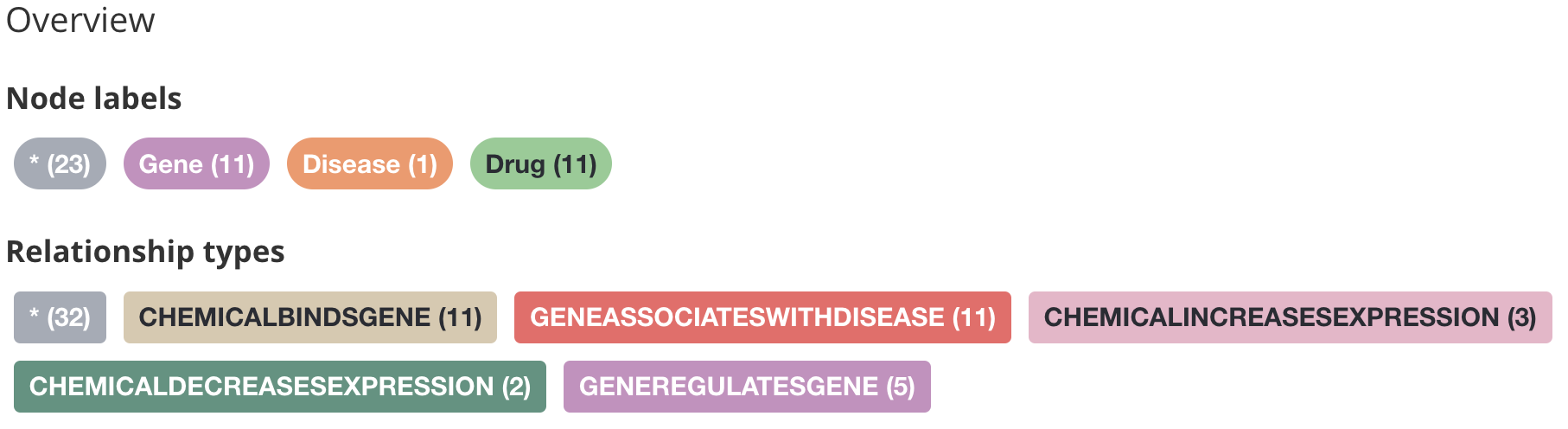}
    \caption{A Small-sized ADKG (Orange Node: AD; Purple Nodes: Genes; Green Nodes: Drugs) \cite{alzheimersknowledgebase}}
    \label{fig:alzkg_small}
\end{figure}

\begin{figure}
    \centering
    \includegraphics[scale=0.04]{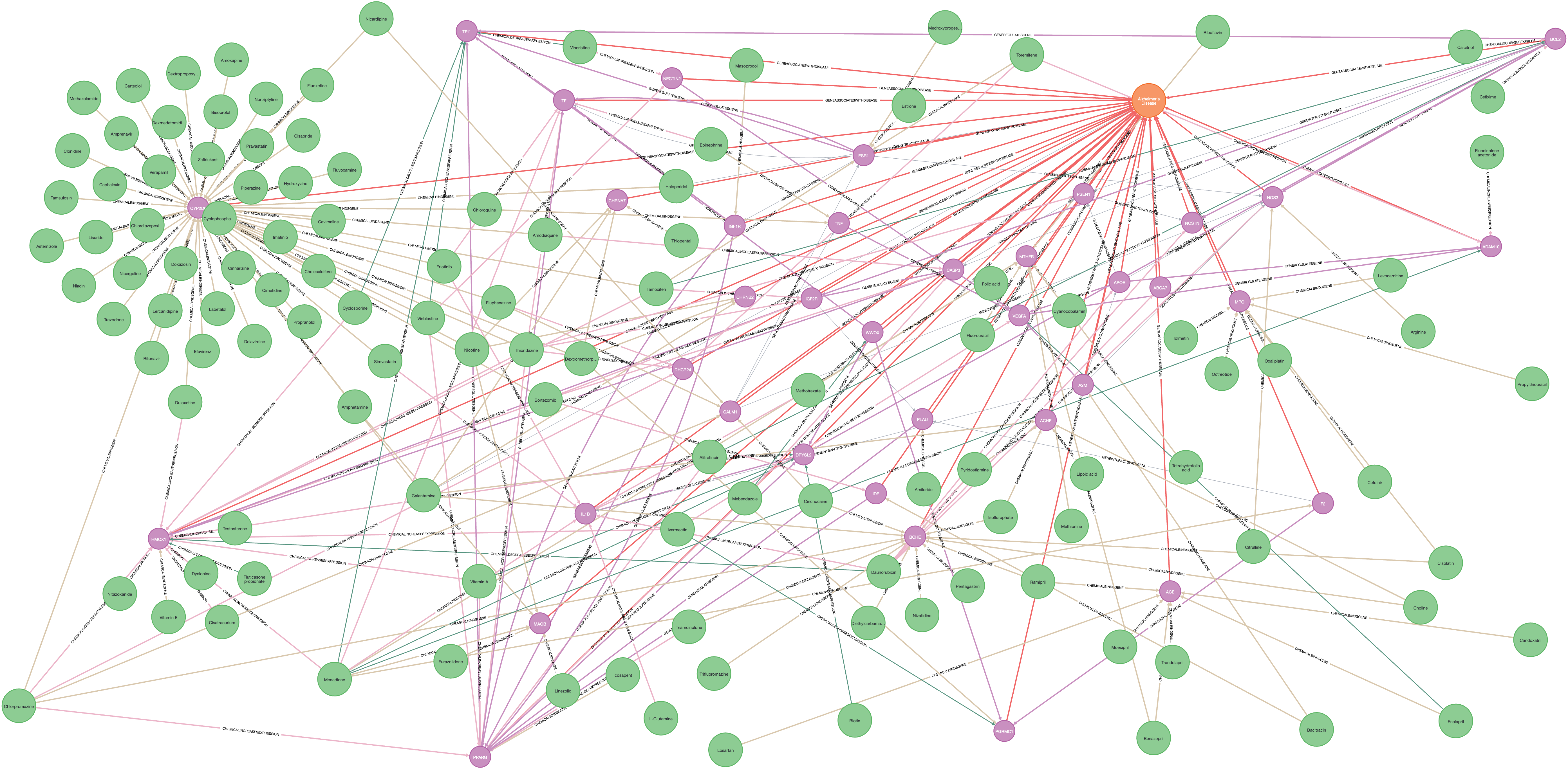}
    \includegraphics[scale=0.2]{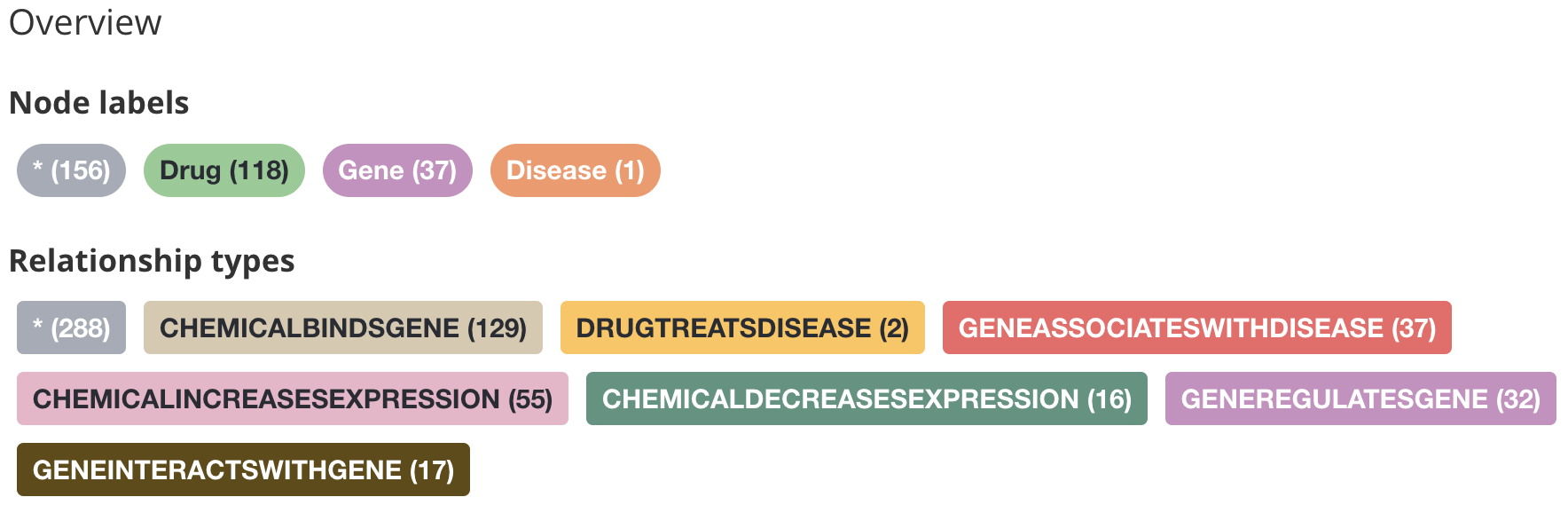}
    \caption{A Medium-sized ADKG (Orange Node: AD; Purple Nodes: Genes; Green Nodes: Drugs) \cite{alzheimersknowledgebase}}
    \label{fig:alzkg_med}
\end{figure}

Within each of the three graphs(see Figure \ref{fig:alzkg_tiny}, Figure \ref{fig:alzkg_small}, and Figure \ref{fig:alzkg_med}), the nodes and their connections are represented by distinct colors and arrow types to convey different biological relationships:

Orange (see Figure~\ref{fig:alzkg_small} and Figure~\ref{fig:alzkg_med}) and Yellow (see Figure~\ref{fig:alzkg_tiny}) nodes represent the disease entity, with AD positioned as the central node, highlighting it as the primary focus of this network.

Purple nodes signify genes, which are implicated in AD through various associations such as genetic risk factors, differential gene expression, or other genetic interactions.

Green nodes denote chemicals, encompassing drugs, vitamins, or other bioactive molecules. These external agents are potential modulators of gene function or disease pathology.

\begin{itemize}
    \item There are five instances of the ``Chemical binds gene" relationship (light purple arrows in Figure~\ref{fig:alzkg_tiny} and coffee arrows in Figure~\ref{fig:alzkg_small} and Figure~\ref{fig:alzkg_med}), where a chemical is shown to interact directly with a gene. This does not necessarily indicate an increase or decrease in gene expression, but rather a physical or functional interaction. For example, one of the edges indicates folic acid, a form of vitamin B that is vital for making DNA and other genetic material, binds the MTHFR gene. MTHFR plays a crucial role in processing amino acids, the building blocks of proteins. Variants of this gene can affect homocysteine levels in the blood. Deficiencies in folic acid are linked to elevated homocysteine levels, which may increase AD risk.
    \item There are six instances of the ``Gene associates with disease" relationship (yellow arrows in Figure~\ref{fig:alzkg_tiny} and red arrows in Figure~\ref{fig:alzkg_small} and Figure~\ref{fig:alzkg_med}), representing genes that have an association with AD. These relationships might represent genetic risk factors, genes involved in the pathology of the disease, or genes that could be potential targets for therapeutic intervention. For instance, the NOS3 gene is associated with AD. It is involved in the generation of nitric oxide, a molecule that aids in blood vessel dilation. Impairment in NOS3 function can affect blood flow in the brain, potentially impacting Alzheimer's disease pathology.
    \item  There are three instances of the
    ``Chemical increases expression"
    relationship (pink arrows), denoting chemicals that are known to upregulate or increase the expression of certain genes. For instance, Vitamin A increases the expression of HMOX1, a gene-encoding enzyme in response to oxidative stress, which is a contributing factor in neuronal damage observed in AD.
    \item There is one instance of the 
    ``Chemical decreases expression"
    relationship (green arrow), indicating a chemical that downregulates or decreases the expression of a gene. Namely, Cyclosporine, an immunosuppressant that may inhibit the formation of amyloid plaques, a hallmark of AD, decreases the expression of TPI1, an enzyme that plays a crucial role in glycolysis, a metabolic pathway that occurs in the cytoplasm of cells.
    \item There is one instance of  ``Gene regulates gene" (purple arrow), suggesting a regulatory interaction between two genes, PPARG and TPI1. For context, PPARG is a gene that codes for a protein that regulates fatty acid storage and glucose metabolism. It is a target for some drugs that might influence Alzheimer’s disease progression.
\end{itemize}   

Figure~\ref{fig:alzkg_small} provides an example of small-sized ADKG with 23 nodes and 32 edges (setting the Cypher limit clause to 15) and figure~\ref{fig:alzkg_med} provides an example of medium-sized ADKG with 156 nodes and 288 edges (setting the Cypher limit clause to 200). In addition to the relationship types described above, the medium-sized ADKG also demonstrates the ``DRUGTREATDISEASE" (gold arrows) and ``GENEINTERACTSWITHGENE" (brown arrows) relationships. As the size of KGs continues to expand, the challenge of comprehending the intricate web of entities and relationships within them becomes daunting for human observers. Consequently, there arises an urgent need for the development of sophisticated computational tools capable of effectively managing these vast KGs.

%\subsubsection{Medium KG}

\bibliographystyle{unsrt}
\bibliography{refs}

%% sample for biography with author's image
% {\color{black!20}\rule{77pt}{77pt}}
\begin{biography}{}{\author{Sisi Shao} is currently a PhD student at the Department of Biostatistics, UCLA Fielding School of Public Health. She received a M.Sc. degree in Financial Engineering from UCLA Anderson School of Management. Her research interest includes artificial intelligence, automated machine learning methods, and statistical methods for large-scale multivariate time series and longitudinal data.}
\end{biography}

%% sample for biography without author's image
\begin{biography}{}{\author{Pedro Henrique Ribeiro}  received a M.S.E. in Bioengineering from the University of Pennsylvania and a B.A. in Computer Science from Oberlin College. He is currently a research data scientist at the Cedars-Sinai Department of Computational Biomedicine. His main research interests are in machine learning and evolutionary algorithms.}
\end{biography}

\begin{biography}{}{\author{Christina Ramirez } received her Ph.D. degree in Statistics/Social Science from the California Institute of Technology, Pasadena, CA, USA, in 1999. She is currently a Professor of Biostatistics at the UCLA Fielding School of Public Health. Her main research interests include HIV pathogenesis, HIV drug resistance mutation/recombination, viral fitness, coreceptor utilization, and high-dimensional data analysis. }
\end{biography}

\begin{biography}{}{\author{Jason H. Moore} is Chair of the Department of Computational Biomedicine at Cedars-Sinai Medical Center where he also serves as Director of the Center for Artificial Intelligence Research and Education. His research focuses on the development and application of artificial intelligence and machine learning methods for the analysis of biomedical and clinical data with the goal of improving health.}
    
\end{biography}

\end{document}